# The Knowledge Layer Model (KLM): Evidence-Visible AI Output, Routing Diagnostics, and Negative Results

Omer Asim Ozturk, Cihan Aydiner

## Abstract

AI answers require more than apparent correctness: reviewers need to inspect their recorded basis. We introduce the Knowledge Layer Model (KLM), an open, implementation-neutral specification for inference records with versioned epistemic assessments, evidence relations and explicit unknowns. We describe SEDIM, the knowledge and evidence architecture of Nage AI OS, as a reference implementation. The specification reference is KLM v0.3, distributed under repository tag v0.3.0. Our empirical material comprises separate internal experiments, not an end-to-end validation. A historical platform-domain routing benchmark reports 137/150 classifications across 15 classes (91.3%; nominal 95% Wilson interval 85.7–94.9%); the numerator is recovered from a rounded summary, and historical example membership remains unverified. A native-model weight-path configuration yielded 0/15 successful answers in its corrected two-source test. A separate capacity-growth comparison is limited by a defective legacy scorer and missing original outputs; a reliable capacity effect could not be established. Lexical and scoped diagnostics each yielded AUC 0.008 for silent-versus-contradicted examples, below the registered minimum of 0.65. Under legacy packing, 118/221 selected cases lost gold-containing text before model input; repaired packing retained it in 221/221 cases. These are text-presence results, not answer-quality results. Python and TypeScript verifiers reportedly agreed on six checks per synthetic fixture; a separate production disclosure was not accepted. Restoration, re-derivation, human-rubric support, boundary isolation and bounded-response evaluations remain proposed. Findings are internal and author-reported, with historical observations distinguished from later reproductions. Evidence visibility does not establish correctness, causal reliance, security or regulatory compliance; independent replication and a complete public audit package remain future work.

## 1. Introduction

Large language model systems increasingly support organizational work in which a fluent answer is not enough. Analysts, auditors, operators and decision-makers need answers to a different set of questions. Which evidence supports this material claim? Which knowledge boundary governed the response? Can that evidence be recovered after an index, model or policy update? What does the system do when evidence is missing, conflicting, inaccessible, or outside its authorized scope?

Answer-only evaluation cannot settle these questions. A response may be coherent, and even factually correct, while giving a qualified reviewer no inspectable basis for judging whether its material claims are supported, appropriately scoped or reproducible. The problem is sharpest in high-accountability settings such as public administration, homeland security, healthcare and finance, and in other rights- and safety-impacting uses. It arises wherever AI output informs consequential analysis, planning or compliance activity. Governance frameworks now ask organizations to document, measure and monitor AI systems across their lifecycle [NIST, 2023; NIST, 2024]. Regulatory regimes attach logging, transparency and human-oversight obligations to higher-risk deployments [European Parliament and Council, 2024]. These instruments specify what an organization must be able to show. They largely leave open how an individual generated claim is to be connected to evidence that a reviewer can inspect and later reconstruct. That implementation-level gap is the subject of this paper.

We call the operational property of making such support inspectable evidence visibility. Evidence visibility is not the same as displaying a citation, logging a retrieval event or presenting a model-generated explanation. A citation can be irrelevant, incomplete, stale, inaccessible, or unrelated to the claim it appears to support. A retrieval log can show that a document was retrieved without showing that it supported the generated claim, or that the same evidence can be recovered later. The central question is therefore not whether a system retrieves or cites sources. It is whether a material output claim can be connected to inspectable, versioned and recoverable evidence under an explicit knowledge boundary.

This paper introduces the Knowledge Layer Model (KLM), a proposed open specification, released under the Apache 2.0 license, for the shape of evidence-visible AI output [Ozturk, 2026]. KLM specifies the information an inference record must carry to make its declared basis inspectable; it does not prescribe how a model produces that information. The record links claims, evidence, procedures and assessments, preserves epistemic statuses and reasoned unknowns, and identifies the schema and formula versions governing its interpretation. Its metacognitive assessment distinguishes declared confidence, grounded confidence and their gap, as detailed in Sections 2.4 and 5.2. The reference implementation is SEDIM, the knowledge and evidence architecture of Nage AI OS. The evaluation draws on multiple historical configurations. The RQ1 routing result comes from the nm/fehm-v3.1 configuration evaluated on 21 April 2026; the claim-assessment, record-verification and native-model experiments use separate configurations. Section 7 identifies the retained references and the configuration information available for each experiment, including identifiers that were not recorded.

Treating attribution, boundary selection, evidence binding and audit records as part of the response path, rather than as annotations added afterward, is what makes these properties testable.

Because KLM is deliberately neutral about implementation, the specification alone cannot establish that an implementation delivers its intended properties. We therefore describe SEDIM and examine the available implementation evidence alongside a proposed evaluation protocol. Historical routing, claim-assessment, verification and native-model results are distinguished from the unexecuted human-rubric support, restoration, re-derivation, boundary-isolation and bounded-response evaluations. The protocol organizes these questions without retroactively imposing one experimental design on all historical runs.

The contribution is deliberately narrower than a claim to solve hallucination, guarantee truth or certify safety. We offer a specification, an implementation and an evaluation approach meant to make AI behavior more inspectable, more open to challenge, and more measurable. We report negative results alongside positive ones. In the historical 21 April 2026 SB-2 benchmark, top-1 platform-domain routing matched the dataset domain in 137 of 150 scored examples across 15 domains. Sections 6.5 and 8.1 define the measure and distinguish retained summary fields from inferred counts. Two design alternatives did not deliver the expected improvement. A weight-path configuration intended to improve route-to-one behavior succeeded in 0 of 15 held-out cases. Route-to-one is an experimental inference path that selects one knowledge unit for a query and applies only its associated weight delta during answer generation. A separate CRATON capacity experiment increased model size from 252M to 642M parameters and reported answerable-item matching rates of 8.0% and 8.8% under a legacy normalized-substring scorer. That scorer was subsequently found to admit spurious matches, and the original comparison could not be rescored because the required outputs were unavailable. These figures are therefore retained as historical measurements, not as validated evidence of a capacity gain. Both the weight-path and capacity experiments concerned the native-model CRATON setting; they examined knowledge delivery and answer generation rather than the platform-domain classification measured by RQ1. These observations motivated further investigation of evidence delivery, evidence binding and measurement instrumentation.

Reporting these failures is a governance choice as well as a scientific one. Capability claims about AI systems are often supported by documentation that describes intended behavior rather than demonstrated behavior. We hold the system to the standard we ask others to apply: a property counts as established only under a stated test condition, success criterion and failure boundary [Aydiner, 2026]. Both authors are affiliated with the organization that develops the evaluated implementation. Section 11 sets out the resulting risk of evaluative bias and the steps taken to mitigate it.

This paper makes four contributions:

1. It defines evidence visibility as a distinct operational property of AI systems, separating claim support, evidence recoverability and knowledge-boundary control from final-answer quality.
2. It introduces the Knowledge Layer Model, a proposed open, implementation-neutral specification for the shape of evidence-visible AI output, and describes a reference implementation that treats route-based attribution and evidence binding as part of the generation path.
3. It specifies separate evaluation procedures for routing, claim support, recoverability, boundary isolation and bounded response. Claim support is assessed at the level of material claims; other evaluations retain their own stated units.
4. It reports positive findings, negative results and limitations, and specifies the reproducibility and audit package needed for independent assessment.

The rest of the paper is organized as follows. Section 2 sets out the conceptual framework, and Section 3 reviews related work. Section 4 states the research questions. Section 5 describes the KLM and its reference implementation. Section 6 defines the evaluation methodology, and Section 7 the experimental setup. Section 8 reports results, and Section 9 the negative results and design alternatives. Section 10 discusses implications. Section 11 addresses limitations, governance and responsible interpretation, and Section 12 concludes. Appendix A specifies the evaluation protocols that are proposed but not yet executed.

## 2. Conceptual Framework

### 2.1 From capability claims to demonstrated properties

The framework starts from a distinction familiar in high-accountability practice: a capability claim is not evidence that the capability has been demonstrated. A statement that an AI system provides attribution, auditability, deletion, isolation or uncertainty awareness must be tested against a defined condition, a success criterion and a documented failure boundary [Aydiner, 2026]. The distinction matters because AI systems can produce plausible outputs, and persuasive documentation, even when the property claimed for them has not been established. It also governs how this paper is written. Each property below is defined together with the condition under which it fails, so that it can be tested rather than merely asserted.

### 2.2 Five operational properties

We distinguish five related but non-interchangeable properties.

**Claim support** concerns whether an evidence artifact actually supports a material claim in the output. It fails when the bound evidence is irrelevant, contradicts the claim, or supports only part of it. A system may display a citation without satisfying this condition.

**Evidence visibility** concerns whether a reviewer can identify and inspect the evidence artifact associated with a material claim, including its source location, version and contextual boundary. It fails when the artifact is absent, inaccessible, or cannot be located at the granularity of the claim it is meant to support.

**Evidence recoverability** concerns whether the evidence path can be reconstructed after generation from the recorded system configuration, knowledge source version, route and retrieval context. It fails when an index, model, corpus or policy update makes the recorded path impossible to restore.

**Knowledge-boundary isolation** concerns whether the system honors the source, tenant, authorization, policy or task boundary governing a query, including under conflicting or adversarial input. It fails when material from outside the authorized boundary is retrieved, disclosed or used to support a claim.

**Authority fidelity** concerns whether the system's role stays appropriately bounded when its output may influence rights-, safety-, mission- or legally consequential decisions. It fails when the system presents unsupported content with unwarranted confidence, or when its output is positioned as a decision rather than as support for one. Evidence-visible output can improve review. It does not transfer accountability from authorized human decision-makers to the system.

The properties are not interchangeable, and a single answer-level score conceals the differences between them. A system can cite exactly the right passage from a document that has since been overwritten, achieving claim support without recoverability. It can enforce its tenant boundaries perfectly while binding claims to evidence that does not support them. It can attach well-supported evidence to an output that is then presented as a recommendation no one is positioned to challenge. Each of these failures calls for a different remedy, and an evaluation that collapses them into one measure of answer quality cannot tell an organization which remedy it needs.

### 2.3 Four distinctions the framework preserves

Four distinctions run through the rest of the paper. Each separates something a system can show from something that requires further evidence.

**Source presence versus evidential support.** Naming a source does not establish that the source supports a particular statement. Citation-evaluation work accordingly scores support at the level of the individual statement rather than the response [Gao et al., 2023].

**Evidential support versus generative reliance.** A passage can support a claim even when the model generated that claim from its parametric memory. Wallat et al. (2025) report that up to 57% of citations in the retrieval-augmented settings they studied lacked citation faithfulness, reflecting post-hoc rationalization rather than reliance on the cited material. Evidence visibility concerns what a reviewer can inspect; it does not by itself establish which evidence influenced generation. Evidence about reliance requires interventions, such as controlled changes to source content, and must be reported separately from evidential support. In our controlled source-change experiment of 20 September 2026, responses reflected the modified source value in all five cases in the tested intervention arm, with no responses retaining the old value. This provides limited behavioral evidence of sensitivity to the modified source under those test conditions. The overall acceptance protocol was not completed, so the 5/5 result is reported as a partial observation in Section 8.6, not as general proof of faithful attribution or successful completion of the protocol.

**Record integrity versus answer truth.** A record can be complete, signed and independently verifiable while describing an answer that is wrong. Integrity establishes that the account of what happened has not been altered. It says nothing about whether what happened was correct.

**Record determinism versus generation determinism.** Individual and institutional users may want the same question to produce the same answer, but settings intended to be deterministic do not by themselves ensure repeatable language-model output. Atil et al. (2024) observed accuracy variations of up to 15% across repeated runs of five models configured for deterministic output. He and Thinking Machines Lab (2025) obtained 80 distinct outputs from 1,000 temperature-zero completions of one prompt; enabling batch-invariant kernels made all 1,000 completions identical in their tested configuration. Generation determinism therefore requires demonstrated control over the inference configuration, rather than an assumption based on temperature or seed settings. The evaluations reported here do not establish such a mode for SEDIM. We instead distinguish record determinism: repeatability of specified verification checks when the record bytes, verifier versions, schema and formula versions, verification keys and other required inputs are held fixed. In the author-run tests reported in Section 8.3, the Python and TypeScript verifiers returned identical verdicts on all six checks for each of two synthetic fixtures. Both implementations were developed within the authors' organization, so this is evidence of agreement on the tested fixtures, not independent replication or a universal verification guarantee. For a user examining a particular answer, reproducible record verification allows the recorded account to be checked without regenerating the answer. It does not establish answer correctness or continued access to every referenced source; evidence restoration is a separate evaluation under Section 6.4.

### 2.4 Explicit unknowns

A further principle follows from these distinctions: an unavailable measurement must remain distinguishable from a measured value. When a required signal cannot be produced, the system must record its absence and a machine-readable reason rather than substitute zero, a default score or fabricated confidence. KLM Specification v0.3, Sections 4.4–4.5, defines this honest-null requirement and requires it to survive the implementation's serialization boundaries. Invalid data must be reported as an error, not silently converted into an unavailable measurement. In the γ contract, an unavailable nullable core score is encoded as null with a corresponding nulls.<signal>.reason. Under klm-gamma/1.1, an unavailable grounded-confidence component instead carries status: unavailable and a reason in grounded_confidence.component_status, without a numeric entry in components. The metacognitive contract distinguishes declared confidence from grounded confidence, preserves the grounded components and their epistemic statuses, and records the declared-minus-grounded gap when both confidence values are available. These are complementary parts of a versioned assessment record, not interchangeable confidence scores. Honest nulls form part of the KLM-0 Declared requirements; describing that requirement does not establish that the evaluated implementation satisfies it. The live-record findings in Section 8.3 are reported separately with their disclosure-profile and verification limits.

### 2.5 Who needs these properties

The properties serve different stakeholders, who ask different questions of the same output.

| Stakeholder | Primary question | Properties most relied on |
|---|---|---|
| Institutional deployers (public agencies, regulated firms, critical-infrastructure operators) | Can we defend a decision informed by this output after the fact? | Recoverability, boundary isolation, authority fidelity |
| Auditors, regulators and oversight bodies | Can we reconstruct what the system did without privileged vendor access? | Recoverability, evidence visibility |
| Analysts and operators | Should I rely on this claim, challenge it, or escalate it? | Claim support, authority fidelity |
| Individual users (professionals, researchers, journalists, students and citizens making consequential personal decisions about health, legal or financial matters) | Can I check what this answer rests on myself, and get the same result when I check again? | Evidence visibility, claim support, record determinism |

Individual users deserve particular attention because they usually lack the audit functions that institutions maintain. Human-factors research shows that people misuse automation by over-relying on plausible output, and disuse it by rejecting sound assistance [Parasuraman and Riley, 1997]. An institution can partly offset both through procedure and review. An individual relying on an AI answer about a medication, a contract or a tax obligation usually cannot. For that user, being able to inspect the basis of an answer directly, and to get the same verdict on a second check, is the practical form that accountability takes. The framework is architecture-neutral for the same reason: no stakeholder should have to accept a vendor's internal component names in order to ask these questions.

### 2.6 From properties to measurement

Table 1 connects each property to the research questions in Section 4, the measures defined in Section 6 and the KLM elements that carry relevant information. RQ1 provides a platform-domain classification diagnostic; it does not inspect claim-level evidence or a delivered record. RQ2 proposes a claim-level support and assessment-coverage evaluation, with earlier automated diagnostics reported separately. Bounded response is intended to operationalize a measurable part of authority fidelity (Appendix A.2; proposed). The remainder depends on who acts on an output and under what review and is addressed through deployment documentation rather than a benchmark.

**Table 1. Mapping of operational properties to research questions, measures and KLM elements.**

| Property | Research question | Measure (Section 6) | Relevant KLM elements |
|---|---|---|---|
| Claim support | RQ2 | Claim-support rate and full rubric distribution (§§6.3, 6.6; proposed human evaluation). Constructed-set scoring diagnostics are reported separately. | Claim identifiers and spans; claim–evidence links; support status and a separate contradiction signal; KLM-2 Grounded. |
| Evidence visibility | RQ1, RQ2 | Platform-domain routing accuracy as an internal routing diagnostic (RQ1); assessment coverage under the proposed claim-level rubric (RQ2). Routing accuracy alone does not measure claim-level evidence visibility. | Inference identifier; evidence references and source metadata; claim-to-source relations; KLM-1 Traceable and KLM-2 Grounded. KLM-4 Attributable adds traceability from reflective findings to the objects assessed. |
| Evidence recoverability | RQ3 | Restoration and re-derivation rates, immediately and after an update (§6.4; proposed). Record-verification results are reported separately. | Evidence identifiers, source versions, content hashes and recorded configuration; lifecycle information where applicable. KLM-5 Verifiable concerns independent record verification, not guaranteed evidence restoration. |
| Knowledge-boundary isolation | RQ4 | Violation rate by class: unauthorized retrieval, exposure, disclosure and incorrect boundary attribution (Appendix A.1; proposed). | Declared request scope; authorization and scope assessments; policy identifiers, versions and provenance; KLM-3 Reflective. These fields document controls and assessments; isolation requires separate testing. |

| Property | Research question | Measure (Section 6) | Relevant KLM elements |
|---|---|---|---|
| Authority fidelity (measurable part) | RQ5 | Appropriate-bounding rate, inappropriate-abstention rate and label consistency (Appendix A.2; proposed). | Versioned epistemic-label rules; source-tension records; metacognitive and governance assessments; KLM-3 Reflective. |
| Cross-cutting: design alternatives | RQ6 | Experiment-specific outcomes for the native-model weight-path, capacity and packing studies (§§9.1–9.4), with their scoring limitations. These are separate from the RQ1 routing benchmark. | No single conformance level. Relevant requirements depend on the record properties examined by each experiment. |

Taken together, these properties move evaluation beyond whether a final response appears correct. They ask whether a qualified reviewer can inspect the basis recorded for an output, assess its support and authorized scope, and determine whether its artifacts remain available after change. The recorded basis is not itself proof of generative reliance; that requires separately scoped intervention evidence.

# 3. Related Work

This paper sits where four bodies of work meet: the evaluation of retrieval-augmented generation, the evaluation of attributed text and citations, standards for recording provenance, and frameworks for AI governance and human oversight. We review each in turn and then state how the KLM differs.

### 3.1 Retrieval-augmented generation and its evaluation

Retrieval-augmented generation combines a model's parametric memory with an external, non-parametric store, which allows knowledge to be updated without retraining and gives outputs a potential basis in identifiable sources [Lewis et al., 2020]. Retrieval does not, however, establish that a generated claim is supported by what was retrieved. A system may retrieve relevant passages and still generate unsupported conclusions, omit attribution for claims that need it, attach citations that do not entail the associated statement, or cite evidence that cannot later be reconstructed. Evaluation work has responded by treating context relevance, faithfulness, answer relevance, correctness and citation quality as separate dimensions. A recent systematic review consolidates these dimensions and documents wide variation in how they are measured in practice [Knollmeyer et al., 2026]. RAGAS operationalizes faithfulness, answer relevance and context relevance as reference-free metrics over retrieved contexts [Es et al., 2024]. Citation-aware retrieval and synthesis systems for scientific literature show the same separation between retrieval quality and evidence-grounded synthesis [Ding et al., 2026]. These approaches measure the quality of a pipeline. They do not specify what record an individual output should carry so that a later reviewer can check it.

### 3.2 Attributed generation and citation evaluation

Work on attributed text generation asks whether a generated statement can be attributed to identified source material, and it typically takes the claim or sentence, not the whole response, as the unit of analysis. ALCE established citation recall and citation precision as statement-level metrics for generated text with citations [Gao et al., 2023]. The choice of unit matters for deployment. A polished response can mix supported, weakly supported and unsupported claims, and an answer-level judgment cannot separate them. Recent work on deep research agents makes the point concrete. When cited content is actually retrieved and checked against the claims, link validity and topical relevance remain high while factual accuracy against the cited source drops sharply, and accuracy falls further as the number of tool calls grows [Onweller et al., 2026].

This literature also shows how hard the judgment is to automate. On AttributionBench, even a fine-tuned GPT-3.5 reaches only about 80% macro-F1 on binary attribution classification, and most errors involve nuanced information the evaluator fails to process [Li et al., 2024]. On CiteME, language models identify the correct source paper for a claim with 4.2–18.5% accuracy, against 69.7% for human evaluators [Press et al., 2024]. CiteEval argues that binary entailment is a poor proxy for citation quality and scores citations against the full retrieval context on a finer scale [Xu et al., 2025]. These findings support the position taken in Section 6: automated evaluators can help with triage, but high-stakes claim-support judgments need a human rubric.

### 3.3 Faithfulness, reliance and reproducibility

A further line of work separates whether a citation supports a statement from whether the model actually relied on it. Wallat et al. (2025) distinguish citation correctness from citation faithfulness and report that up to 57% of citations in the settings they studied lacked faithfulness, the result of post-hoc rationalization rather than reliance. Agent systems widen the problem. A survey of evidence tracing and execution provenance in language-model agents treats execution provenance as a typed graph of an agent's run, and evidence tracing as its projection onto evidence-support relations, linking claim support to tool use, memory lineage, audit and recovery [Wang et al., 2026].

Reproducibility is a separate obstacle to after-the-fact review. Language-model outputs vary across repeated runs even under settings intended to be deterministic. Atil et al. (2024) report accuracy variations of up to 15% across runs of five models configured for deterministic output. He and Thinking Machines Lab (2025) traced most temperature-zero variation to inference kernels that are not invariant to batch size, and showed that batch-invariant kernels can restore bitwise reproducibility at some cost in speed. For governance, the implication is that re-running a model is not a dependable way to audit what it did on an earlier occasion. An audit has to rest on a record made at the time. This motivates the distinction between record determinism and generation determinism in Section 2.3.

### 3.4 Provenance and output-record standards

Several existing standards record where an artifact came from, and the KLM has to be placed among them. The W3C PROV family defines a general data model and ontology for provenance, expressed as entities, activities and agents, that can be specialized for particular domains [Lebo et al., 2013]. The C2PA Content Credentials specification binds cryptographically signed manifests to media assets, so that a validator can confirm an asset's recorded origin and edit history and detect tampering [C2PA, 2026]. The OpenTelemetry semantic conventions for generative AI standardize how model calls, agent invocations, tool executions and evaluation results are named in operational telemetry, so that traces from different frameworks can be read by any compatible backend [OpenTelemetry, 2026].

Each of these standards answers a different question from the one this paper addresses. PROV can express that an output was derived from a document, but it leaves open whether the document supports a particular claim in the output. C2PA supports validation of signed provenance assertions and their binding to an asset under a defined trust model. It does not establish the factual truth of the asset's content. It is not designed to describe whether the statements in a text are grounded in evidence. OpenTelemetry records which model was called, with what parameters and at what cost, and is built for system operators rather than for reviewers of individual claims. None of the three records an output's epistemic state: how far its claims are grounded, whether its sources conflict, whether its evidence is current, and what could not be assessed.

**Table 2. Comparison of provenance and output-record standards.**

| Standard | Unit of record | What it establishes | What it does not establish |
|---|---|---|---|

| Standard | Unit of record | What it establishes | What it does not establish |
|---|---|---|---|
| W3C PROV-O [Lebo et al., 2013] | Entity, activity, agent | Derivation and attribution relations in a general provenance graph | Whether a source supports a specific claim; epistemic state |
| C2PA Content Credentials [C2PA, 2026] | Media asset and signed manifest | Origin, edit history and tamper evidence of an asset | Whether statements in the content are grounded or true |
| OpenTelemetry GenAI conventions [OpenTelemetry, 2026] | Span, metric and event for a model or agent operation | Which model, tool and parameters were used; latency and token usage | Claim-level evidence support; recoverable evidence paths |
| Model Cards and Datasheets [Mitchell et al., 2019; Gebru et al., 2021] | Model or dataset | Intended use, evaluation conditions, known limitations | Properties of an individual output |
| Knowledge Layer Model (this paper) [Ozturk, 2026] | Individual AI output | A versioned attestation graph linking claims, evidence, procedures and assessments, with explicit epistemic statuses and reasoned unknowns; independent record verifiability is an additional KLM-5 requirement. | Answer truth; generative reliance; legal compliance |

The KLM is designed to complement these standards rather than replace them. KLM Specification v0.3, as distributed under repository tag v0.3.0, does not define normative mappings to PROV-O, C2PA or the OpenTelemetry GenAI semantic conventions. We therefore treat formal mappings, exchange profiles and tests of information preservation across these representations as future work. Conceptual compatibility alone does not establish interoperability, and no implemented or validated integration with these standards is claimed here.

### 3.5 Documentation and AI management frameworks

Documentation and management frameworks work at the level of the system or the organization. Model Cards report a model's intended use, evaluation conditions and performance across groups [Mitchell et al., 2019], and Datasheets for Datasets do the same for training data [Gebru et al., 2021]. The NIST AI Risk Management Framework organizes risk management across the AI lifecycle through its Govern, Map, Measure and Manage functions [NIST, 2023]. Its Generative AI Profile identifies risks and recommended actions related to documentation, measurement, testing, monitoring, content provenance and human oversight [NIST, 2024]. ISO/IEC 42001 specifies requirements for establishing, maintaining and continually improving an AI management system within an organization [ISO/IEC, 2023]. These instruments tell an organization what to document, measure and manage. They do not specify a technical mechanism for binding a single generated claim to evidence that can be recovered later. In practice, an organization that has adopted the NIST AI RMF or achieved ISO/IEC 42001 certification may still be unable to answer, for a given output, what its material claims rested on.

### 3.6 Human oversight, security and regulation

Human-factors research has long shown that the reliability of automated advice is not a property of the system alone. Parasuraman and Riley distinguish use, misuse, disuse and abuse of automation. Operators may over-rely on plausible output or reject sound assistance, and system design shapes which of these happens [Parasuraman and Riley, 1997]. A review of 76 empirical studies of human-autonomy teaming identifies agent characteristics, task characteristics, training and communication as key influences on team outcomes, and points to trust and shared awareness as recurring concerns [O'Neill et al., 2022]. Joint guidance from the NSA, CISA, the FBI and Five Eyes partners recommends practices for deploying and operating externally developed AI systems securely, especially in high-threat, high-value environments [NSA AISC et al., 2024].

Regulation increasingly turns these concerns into obligations. The EU AI Act requires high-risk AI systems to support automatic record-keeping, to give deployers enough transparency to interpret outputs, and to be designed for effective human oversight [European Parliament and Council, 2024]. In the United States, OMB Memorandum M-25-21 directs federal agencies to apply minimum risk-management practices to high-impact AI and to maintain public transparency about its use [OMB, 2025]. These obligations specify what must be possible, such as reconstructing events, interpreting outputs and enabling oversight. They leave the technical means to implementers. Governance-grade AI frames the practitioner's task as establishing, for a specific mission, whether a system demonstrably supports attribution, auditability and scoped information control, rather than relying on design descriptions or vendor claims [Aydiner, 2026].

### 3.7 Positioning

Taken together, this work operates at three levels. Organizational frameworks (NIST AI RMF, ISO/IEC 42001) govern how an institution manages AI. System documentation (Model Cards, Datasheets) describes a model or dataset. Evaluation benchmarks (ALCE, RAGAS, AttributionBench, CiteEval) measure how well a pipeline performs across a test set. What remains largely unstandardized is the individual output: a portable record, attached to each answer, that states what its claims rest on, what is uncertain or contested, and what could not be assessed, in a form a reviewer can verify without access to the vendor's internal systems.

The KLM addresses that output level. Our contribution differs from citation-generation and retrieval-evaluation work in four respects. It proposes an open, implementation-neutral output specification. It distinguishes route classification, claim support and knowledge-boundary isolation and specifies separate tests for them. It treats evidence recoverability and explicit unknowns as record requirements, while distinguishing those requirements from demonstrated implementation capabilities. It also reports negative design results and measurement defects as part of the empirical record. It does not claim that these mechanisms establish factual correctness. It asks how recorded assertions can be inspected, challenged and tested under defined conditions.

## 4. Research Questions

The research questions turn the five properties of Section 2 into measurable questions (Table 1). Each is paired with a measure in Section 6 and an evaluation status in Table 3. Historical diagnostic results are distinguished from the proposed human-rubric evaluation for RQ2, restoration and re-derivation for RQ3, and the RQ4–RQ5 protocols in Appendix A. A proposed evaluation is not an empirical finding.

**RQ1 — Platform-domain classification.** In the historical closed-set routing benchmark, how often does the largest averaged routing weight match the dataset-derived domain label across the router's 15 classes, and what domain-level results survive in the retained summary?

**RQ2 — Claim support.** To what extent do the evidence artifacts bound to material claims support those claims under a predefined rubric, and what share of material claims have no assessable bound evidence at all?

**RQ3 — Evidence recoverability.** Can a reviewer, using only the recorded identifiers and versions, restore the evidence artifacts bound to an output, and can the evidence path be re-derived under the recorded configuration, both immediately and after a subsequent index, model or policy update?

**RQ4 — Knowledge-boundary isolation.** Does the system keep to its defined source, tenant, authorization and policy boundaries when presented with cross-boundary, conflicting or adversarial inputs, and which classes of violation, if any, occur?

**RQ5 — Bounded response.** When evidence is missing, insufficient, inaccessible, materially conflicting or outside the authorized scope, does the system constrain its output instead of presenting unsupported content with unwarranted confidence? And when sufficient evidence is available, does it avoid withholding or over-qualifying a supportable answer?

RQ6 — Design alternatives. What experiment-specific outcomes and measurement limitations are reported for the native-model weight-path configuration, capacity-growth comparison and evidence-packing alternatives, and what further tests do they motivate? These are separate from the RQ1 classification benchmark.

The second clause of RQ5 is deliberate. A system that declines every difficult question would satisfy the first clause perfectly and be useless in practice. Bounded response is meaningful only when both failure directions are measured.

**Table 3. Evaluation status of the research questions.**

| RQ | Primary measure | Evaluation status |
|---|---|---|
| RQ1 | Top-1 platform-domain routing accuracy | Historical internal SB-2 run, 21 April 2026. The numerator 137/150 is recovered from the retained summary. Individual predictions and a verified historical split manifest are unavailable. The author reports training and validation files in a private archive, but historical membership has not been verified. The summary and provenance records have not been publicly released. |
| RQ2 | Claim support rate and full rubric distribution | The human-rubric evaluation specified in §§6.3 and 6.6 is proposed; no completed evaluation under that rubric is reported. Historical support-scoring and contradiction-detection diagnostics on constructed sets are reported separately and do not constitute execution of this protocol. |
| RQ3 | Restoration rate and re-derivation rate | Proposed (§6.4). No completed immediate or post-update restoration and re-derivation evaluation is reported. Record-verification and source-hash consistency checks are reported separately; they do not establish evidence recoverability. |
| RQ4 | Violation rate by violation class | Proposed (Appendix A.1). No completed evaluation under this protocol or completed live two-tenant isolation acceptance result is reported. |
| RQ5 | Appropriate-bounding rate; inappropriate-abstention rate; label consistency | Proposed (Appendix A.2). No completed evaluation reporting these measures is included. |

| RQ | Primary measure | Evaluation status |
|---|---|---|
| RQ6 | Experiment-specific outcomes for native-model weight-path, capacity and packing alternatives | Historical internal CRATON experiments, author-reported. The weight-path test reports 0/15. The capacity comparison used a legacy scorer subsequently found to admit spurious matches and could not be rescored from the required outputs. Packing results concern evidence presence in the final input, not answer correctness. These experiments are separate from the RQ1 routing benchmark. |

Results reported in Section 8 are identified by who conducted the evaluation and under what conditions. The findings reported here are internal and author-reported; none is presented as independently replicated. Blinded scoring, where used, must be identified separately and does not by itself establish evaluator independence. Historical results and separately dated reproductions are distinguished from one another. An evaluation marked "proposed" is specified as a protocol but is not reported as completed; it must not be cited as an empirical finding.

# 5. The Knowledge Layer Model

This section separates the KLM specification from its reference implementation. Sections 5.1–5.4 describe the record model, epistemic labels, conformance requirements and stewardship arrangements. Sections 5.5–5.10 describe SEDIM and the evidence available for its implementation status. The specification can be adopted without adopting SEDIM; describing a requirement does not establish that the evaluated implementation satisfies it.

### 5.1 Scope and design principles

KLM Specification v0.3 defines the information an implementation must emit and preserve to make an inference inspectable through a structured record. It does not prescribe a model architecture, retrieval algorithm or training method. Its descriptive model separates three dimensions: knowledge origin, functional role and reflective assessment. The six descriptive layers, L0–L5, occupy positions within these dimensions; they are not a sequence of mutually exclusive processing stages and must not be confused with the six cumulative conformance levels.

The record distinguishes claims from evidence, source presence from claim support, and execution attribution from causal contribution. Every signal carries an epistemic status: measured, heuristic, synthesized or unavailable. An unavailable required signal carries a reasoned null; invalid data is reported as an error. Formula identifiers preserve the meaning of computations across implementations and over time. These requirements make recorded assertions open to checking without treating them as guarantees of answer correctness.

### 5.2 The KLM record and epistemic labels

The unit of conformance is an inference. Its attestation record is a graph linking knowledge units, evidence objects, claims, output spans, procedures and reflective assessments. Typed relations distinguish support, contradiction, procedural activation, assessment and derivation. The record also carries model and configuration references, request scope, policy provenance and explicit unknowns. Span offsets declare their counting unit and identify the output text they index.

The metacognitive assessment separates declared confidence from grounded confidence. Grounded confidence preserves its components, their epistemic statuses and its declared semantics; the confidence gap records declared confidence minus grounded confidence when both are available. An epistemic label is a versioned projection of these signals, not a replacement for them.

Table 4 summarizes the registered klm-label/2.0 projection, which requires klm-gamma/1.1. It describes a specification option, not the label formula used by every historical record or by the live implementation. Records declaring klm-label/1.0 retain that version's rules.

**Table 4. Ordered trigger rules for the registered klm-label/2.0 projection.**

| Order | Label | Trigger |
|---|---|---|
| 1 | SPECULATIVE | Evidence < 0.05. |
| 2 | CONTESTED | Coherence < 0.40 and evidence ≥ 0.20. |
| 3 | OUTDATED | Freshness < 0.30. |
| 4 | STABLE | Grounded confidence ≥ 0.65, evidence ≥ 0.60, coherence ≥ 0.72 and attributed mass ≥ 0.65. |
| 5 | PROBABLE | Either evidence ≥ 0.35 and declared confidence ≥ 0.45, or grounded confidence ≥ 0.55. |
| 6 | UNCERTAIN | No preceding applicable rule matches. |

Rules are evaluated in order, and the first matching rule determines the label. Here, evidence, coherence and freshness denote the corresponding γ score fields; grounded confidence denotes grounded_confidence.value, and attributed mass denotes attributed_mass. A condition requiring an unavailable input is skipped rather than evaluated as zero. For PROBABLE, each alternative is evaluated separately. Threshold changes require a new formula identifier. The labels summarize the declared computation; they do not independently establish source entailment, factual correctness or appropriate reliance.

### 5.3 Cumulative conformance levels

The specification defines six cumulative conformance levels. Each level adds requirements to the preceding ones. Table 5 summarizes the required properties and the limits of the resulting claim; it does not assign an operating level to SEDIM.

**Table 5. KLM conformance levels and limits of interpretation.**

| Level | Required property | What it does not establish |
|---|---|---|
| KLM-0 Declared | A structured inference record, explicit epistemic statuses and reasoned unknowns, without fabricated signals. | That the answer is supported, correct or independently verified. |
| KLM-1 Traceable | Replayable records of generative inputs and key events, including model version, retrieval trace, knowledge-unit identifiers, loaded procedures and output spans. | That retrieved material supports a claim or influenced generation. |
| KLM-2 Grounded | Mechanically traceable claim–evidence and procedure–execution links, together with the required evidence and attribution signals. | That an automated support measure establishes entailment, or that linked evidence caused the answer. |
| KLM-3 Reflective | Separate metacognitive and governance traces, including split confidence, the confidence gap and policy provenance. | Independent organizational review, sound policy choices or regulatory compliance. |

| Level | Required property | What it does not establish |
|---|---|---|
| KLM-4 Attributable | Reflective findings traceable through graph relations to the generative objects they assess, with execution attribution for governed behaviors. | A complete causal explanation of model-internal reasoning. |
| KLM-5 Verifiable | Record properties checkable by a third party through the specified integrity and verification mechanisms. | Answer truth, continued evidence availability or an omission-evident audit trail without the additional completeness mechanisms. |

A conformance claim identifies the record profile, schema and formula versions, test version and evaluating party, including whether that party is independent of the implementer. Passing sample records demonstrates a capability under those conditions; claiming an operating level requires evidence that the implementation's emitted records consistently meet that level. Historical records are evaluated under their declared versions, rather than retroactively under later requirements.

### 5.4 Stewardship, versioning and licensing

The KLM specification and reference toolchain are published in the public omasim/KLM repository under the Apache-2.0 license [Ozturk, 2026]. This paper references Specification v0.3 as distributed under repository tag v0.3.0. Open licensing permits reuse under the license terms; it does not establish independent governance or certify an implementation's conformance.

Nage AI stewards the specification and develops SEDIM, creating a potential conflict between the interests of the specification and those of its reference implementation. The public contribution policy defines an amendment process through GitHub issues: contributors describe the problem, propose a change and provide a concrete example; discussion and the maintainer's decision are recorded publicly, and accepted amendments enter a subsequent specification version and changelog. Under the documented v0.x process, decision authority remains with the maintainer. Public participation therefore does not amount to shared or independent stewardship.

The documented safeguards comprise publicly reviewable amendment proposals and decisions, publicly runnable reference validators and conformance vectors, and version-specific conformance claims. Specification versions, toolchain releases, record schemas and formula identifiers serve different purposes and must be identified separately. Historical records are checked under their declared schema and formula versions; later requirements do not retroactively redefine an earlier record's claim. A published conformance claim must also identify the test version, the evaluating party and whether that party is independent of the implementer.

These arrangements make changes and conformance claims open to scrutiny, but they do not eliminate the steward's conflict of interest or establish that the reference tests cover every relevant failure. Implementations developed outside the steward's organization, external evaluation and broader participation in decision-making remain priorities for future work. The implementation status reported in this paper is therefore assessed separately from the specification's requirements.

### 5.5 Request processing and evidence recording

Figure 1 organizes the intended SEDIM request path into five stages linked through a request identifier. First, identity and policy handling establish the declared tenant, access scope and task constraints. Second, knowledge selection identifies candidate knowledge units or platform modules, distinguishing evidence supplied through context from model-side components. Third, for context-based evidence, packing determines the token sequence presented to the model within the context budget. Selection and final-input presence are recorded separately, including truncation information where available. Fourth, the model generates an answer, and claim-support assessments are associated with their method versions and epistemic statuses. Fifth, response delivery and audit-record handling connect the output to the available assessment, integrity and verification information, subject to the disclosure profile.

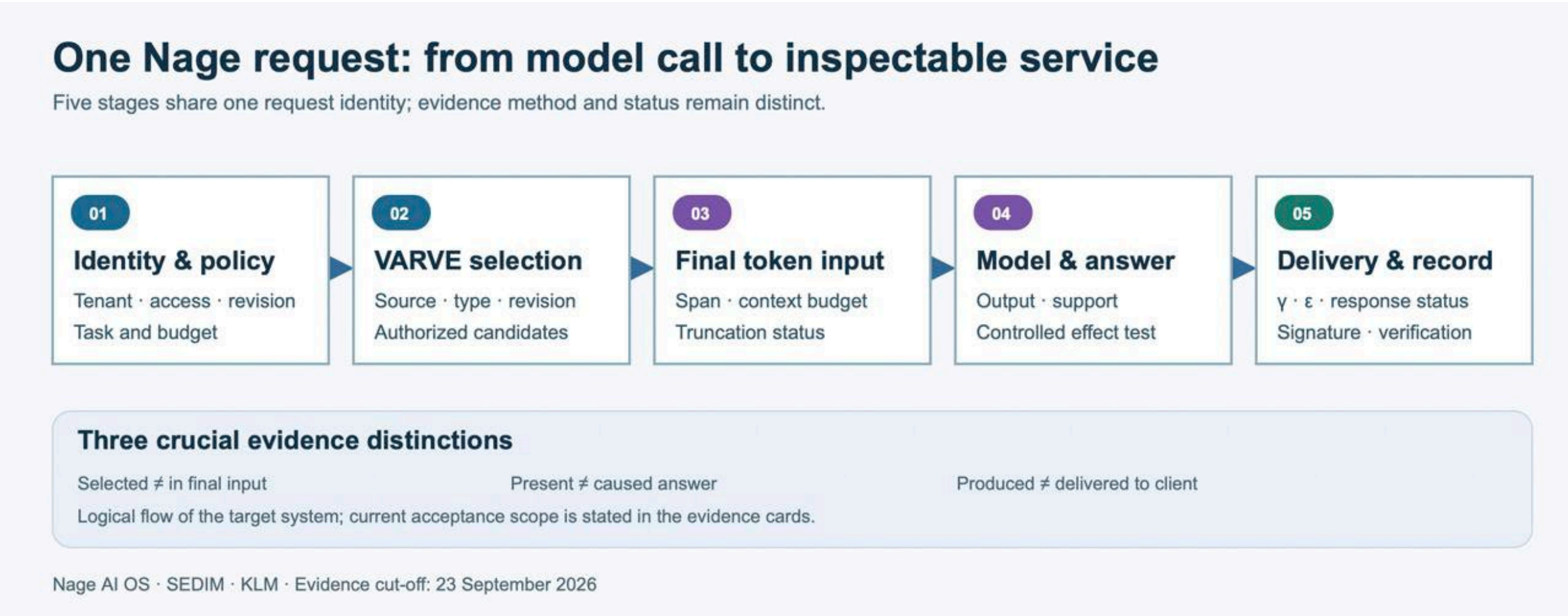

Figure 1. Logical request path and evidence-recording stages in SEDIM. The five stages are correlated through a request identifier. Knowledge selection, final-input presence, behavioral sensitivity and client delivery require separate evidence. The diagram describes the architecture; Table 6 reports component status and evaluation limits.

This is a logical description of the architecture, not evidence that every stage has passed end-to-end acceptance. Table 6 states the implementation status and available evidence for the components. A shared request identifier supports correlation across stages; it does not by itself demonstrate authorization enforcement, complete logging or successful client delivery.

Three distinctions govern the evaluation. Selection does not establish that evidence reached the final model input. Presence in that input does not establish that the model relied on it. Generation of an answer does not establish that the client received the answer or its associated record. Controlled source-change tests investigate behavioral sensitivity separately from ordinary request processing; they are evaluation interventions, not a mandatory stage in every request.

**5.6 Knowledge boundaries and evidence admission**

A knowledge boundary defines which sources, knowledge units and operations are permitted for a request under its tenant, authorization and policy context. In KLM Specification v0.3, knowledge-unit identity and origin are distinguished from version, scope and lifecycle metadata, some of which are optional (§2.1, §10). Recording a scope or policy reference makes the declared boundary inspectable; it does not establish that the implementation enforced it. Unavailable metadata and untested controls must therefore remain distinguishable from demonstrated boundary enforcement.

The reference architecture distinguishes three ways of supplying knowledge. Customer documents can be retrieved and supplied as dynamic context without loading customer-specific weight deltas. Platform modules are model-side low-rank components associated with the platform’s predefined domains; the historical RQ1 benchmark evaluates domain classification within that fixed roster. Neither path establishes a customer-specific parametric realization. That third setting—customer knowledge represented in separately learned weights—requires its own evidence for training, artifact compatibility, loading, application during inference, task benefit, tenant isolation and removal. This paper does not claim that this acceptance chain has been completed.

These settings require separate evidence. A document changing an answer does not demonstrate learning into weights; loading a module does not demonstrate that it contributed to generation. Likewise, restricting retrieved context does not remove information already present in the base model's parameters. The boundary-isolation protocol in Appendix A.1 separately tests unauthorized retrieval, exposure in the final model input, disclosure in the output and incorrect boundary attribution. No completed evaluation under that protocol is reported here.

### 5.7 Domain routing and routing weights

The historical RQ1 configuration uses a supervised router to assign weights across 15 platform-domain classes associated with a frozen Qwen3-8B base. The retained benchmark summary identifies the router as STRIA v3.1 and reports 2,400 training steps, an entropy-regularization weight of 0.30 and 75 dynamic samples per class. These are configuration references, not a retained manifest of the training examples. The evaluated class roster belongs to the nm/fehm-v3.1 configuration of 21 April 2026 and does not represent every module registered across the wider platform.

For each scored input, the benchmark averages the router's weight vectors across token positions and routing blocks. The class with the largest averaged weight is compared with the input's dataset-derived domain label. This produces the closed-set platform-domain classification measure defined in Section 6.5. The benchmark scores neither a generated answer nor the route reported in a delivered attestation record.

Routing weights describe the router's allocation across the configured modules. They do not quantify the factual contribution of each module to an answer, establish claim support or demonstrate causal reliance. The RQ1 diagnostic must therefore remain separate from customer-document retrieval, final-input evidence presence and controlled tests of source influence. Where routing information is included in an audit record, its method and configuration must be identified without presenting the weights as a causal explanation.

### 5.8 Claim–evidence binding and final-input presence

KLM represents claim–evidence binding through typed relations between evidence objects and identified claims or output spans. A recorded link makes the asserted relationship inspectable; it does not establish that the evidence supports the claim or influenced generation. The architecture therefore distinguishes evidence selection, presence in the final model input, claim support and behavioral sensitivity to source changes. Retrieval relevance measures query–source similarity, final-input checks establish text presence under a stated matching rule, and controlled interventions investigate source influence. None of these observations substitutes for the others.

The reference implementation reports a versioned lexical diagnostic, klm-support-lexical/0.1. According to the implementing author's review dated 27 October 2026, this figure appears as a supplementary field within the grounded block and contributes to the context-provenance share; it does not enter the computation of grounded_confidence.value or the epistemic label. Its recorded status of measured describes a mechanical computation, not a validated measurement of entailment. The figure must therefore be interpreted within its lexical-overlap scope. Section 9.3 states the implementation-status evidence and its limits.

KLM Specification v0.3 requires contradiction findings to use a separate, versioned detection signal rather than infer contradiction from a support score alone (§2.2). The dated implementation review reports that the mechanical contradiction detector had not been connected to the live $\gamma$ signal. Its constructed-set and live-answer evaluations are reported separately in Section 8.2. Neither a high lexical score nor the absence of a contradiction flag establishes evidential support.

### 5.9 Records, disclosure and verification

The reference architecture links an output to a stored audit record through an identifier. Verification must distinguish the source record from a selectively disclosed view of it. Where an Ed25519 signature covers a declared record hash, checking the signature authenticates that hash under the verification key; establishing content integrity additionally requires reconstructing the hashed content. A disclosure view that omits required fields may therefore permit signature verification without permitting reconstruction of the signed record. Verification results must state which object and disclosure profile were checked.

The historical Python and TypeScript verifiers described in this paper examine six properties: content integrity, signature authenticity, the declared-minus-grounded confidence relationship, honest-null discipline, the composition manifest and claim-support recomputation. The last check requires the relevant claim and evidence text and reproduces the declared scoring method rather than establishing entailment. Both verifiers were developed within the authors' organization. Agreement between them is reported for the tested records and does not imply that every check passed.

Under the strict acceptance profile, all five required checks must return PASS; a FAIL on an additional conditional check also prevents acceptance. NOT MEASURED remains distinct from FAIL, but neither satisfies a required check. Canonicalization identifiers, verifier versions and verification inputs must be specified. Historical sha256-canonical/* formats must not be treated as interchangeable with the klm-canonical/1 format defined by Specification v0.3. The historical verifiers are also distinct from the klm-conformance reference toolchain.

Section 8.3 reports sample-record acceptance separately from the production operator-view result, which was not accepted. Chunk-hash consistency and Merkle-root checks are likewise separate results: they do not establish answer correctness, continued source availability or an omission-evident audit trail.

### 5.10 Implementation status and known limitations

Table 6 distinguishes implementation availability, reported test results and unresolved acceptance requirements. "Implemented" means that code or a retained artifact exists; it does not imply successful operation under every configuration. "Tested" refers only to the stated population and conditions. Live-use statements are attributed to dated implementation reports rather than treated as independently verified operating guarantees. "Protocol only" denotes an evaluation specified but not reported as completed. "Acceptance incomplete" denotes an implementation claim whose required evidence chain has not been established.

The table does not assign a KLM conformance level to SEDIM. Component availability, agreement between verifiers and successful sample records do not establish that every production record satisfies a conformance profile. Historical measurements, later reproductions and dated operational reports retain their separate scopes.

**Table 6. Implementation status, reported evidence and remaining limitations.**

| Component | Status | Reported evidence and limits |
|---|---|---|
| Platform-domain router | Historical internal test | 137/150 scored classifications in the 21 April 2026 closed-set benchmark. This measures agreement with dataset-domain labels, not source attribution or delivered-record accuracy (§8.1). |
| Evidence retrieval and packing | Implemented; tested in the native-model setting | Historical substring-based visibility: 118/221 selected cases lost before the model; repaired packing retained 221/221. The separate 2 October word-bounded rerun reports 130/219 lost and 219/219 retained after repair. These are text-presence results, not answer-correctness results (§9.4). |

| Component | Status | Reported evidence and limits |
|---|---|---|
| Lexical claim-assessment diagnostic | Implemented; live use author-reported | Lexical and scoped variants each report K3 AUC 0.008 for 12 silent versus 64 contradicted examples. The dated implementation report describes continued use of klm-support-lexical/0.1; its recorded measured status does not establish entailment (§§8.2, 9.3). |
| Mechanical contradiction detector | Implemented and evaluated; live γ integration not established | The separate 25 October holdout reproduction reports detection on 52/59 contradicted items. In the live-answer diagnostic, D1 detects 19/59 INVENTED-labelled answers and flags 34/137 CLEAN-labelled answers; the primary D2 detects 33/59 and flags 46/137. Neither variant met its acceptance criteria. The dated implementation report states that the detector was not connected to live γ (§§8.2, 9.3). |
| Record signing | Implemented; production example examined | Ed25519 authenticity passed on the reported production operator view, but that view did not satisfy the complete acceptance profile. Signature verification is separate from content-integrity verification (§8.3). |
| Python and TypeScript verifiers | Implemented by the authors' organization; agreement reported | The test-keyed sample was accepted. One production operator view was not accepted: 1 PASS, 3 FAIL and 2 NOT MEASURED. Matching verdicts are agreement, not successful acceptance or independent replication (§8.3). |
| Chunk hashes and Merkle root | Historical consistency checks | No hash mismatches across 7,230 chunks; the reported root was unsigned and unannounced. This does not establish an externally witnessed or omission-evident audit trail (§8.3). |
| Final-input evidence probe | One retained live observation | One probe reports a target span at token positions 945–950 within 1,264 input tokens, without truncation. This is not a population-level result or evidence of model reliance (§8.6). |
| Tenant-boundary isolation | Protocol only | Appendix A.1 specifies the evaluation. No completed live two-tenant isolation acceptance result is reported. |
| Bounded response | Protocol only | Appendix A.2 specifies appropriate bounding, inappropriate abstention and label consistency; completed results are not reported. |
| Withdrawal and deletion | Serving-path removal reported; durable erasure not established | The reported removal scope does not establish deletion across every store, replica or derived artifact. Handling of retained records and withdrawn evidence requires explicit disclosure (§11.5). |
| Customer-specific parametric realization | Acceptance incomplete | No completed evidence chain establishes customer-specific training, compatible artifacts, loading, inference-time application, task benefit, tenant isolation and removal (§5.6). |

### 5.11 What the KLM and its implementation are not

The KLM is not a certificate, and conformance is not conferred by its steward. It is not a compliance instrument: a conformant record can support an organization's own assessment under the NIST AI RMF, the EU AI Act or OMB M-25-21, but it does not establish compliance with any of them. It is not an explanation of model-internal reasoning. It is not a guarantee that an output is correct, complete or current. And it is not exclusive to one vendor. The same holds for the reference implementation, which in addition does not remove information already absorbed into a pretrained model's parameters, and does not secure against attack classes outside the tested threat model (Section 11).

# 6. Evaluation Methodology

This section defines the measures and proposed evaluation procedures. A method definition is not a report of completed work. Table 3 distinguishes historical diagnostics from the proposed human-rubric evaluation for RQ2, restoration and re-derivation for RQ3, and the RQ4–RQ5 protocols in Appendix A. Historical runs retain their original populations, scoring methods and provenance limits; they are not presented as execution of every procedure below.

### 6.1 Principles

The proposed evaluation follows four principles. First, the human support evaluation uses the material claim as its unit; routing inputs, evidence paths, records and native-model questions retain their own experiment-specific units. Second, future evaluations must fix their measures, criteria and analysis before examining outcomes. Historical criteria are identified only where dated records support predeclaration, and later scoring corrections are disclosed as post hoc (Section 7.5). Third, distinct properties are reported separately rather than merged into a composite score. Fourth, failed and unassessable cases remain visible in the relevant population accounting; missing evidence is neither silently excluded nor substituted with an invented measurement.

### 6.2 Material claims and segmentation

A material claim is a statement in an output whose truth, scope or qualification could reasonably affect a user's interpretation, decision or action, or a reviewer's assessment of the response. Purely transitional statements, restatements of the question and procedural hedges without factual content are excluded.

Under the proposed human-evaluation protocol, annotators would segment outputs into material claims using a written guideline. The guideline must specify how compound statements, qualifications and overlapping claims are handled. If an automated segmenter proposes candidate claims, human annotators would accept, split, merge or reject them. The evaluation report must identify the segmenter's model, version, prompt and relevant settings, together with the number of proposals and the counts and proportions of human revisions.

A subset of outputs must be selected for independent segmentation by two annotators before agreement is assessed. Its size and selection procedure, the annotators' training and the rules for matching claim spans must be specified before scoring. Agreement is assessed before adjudication; disagreements are then resolved under a documented procedure, preserving both original annotations and the adjudicated segmentation. No completed segmentation study or segmentation-reliability estimate is reported in this paper.

### 6.3 Claim support (RQ2)

For each material claim $c_i$, the evaluator records whether a bound evidence artifact is present, whether it is accessible, and, if it is, a support score on the rubric in Section 6.6: 2 (fully supported), 1 (partially supported), 0 (unsupported), or U (unassessable). Claims with no bound artifact are recorded as missing binding, which is distinct from U.

Let $N$ be the number of evaluated material claims, $N_2$ the number scored 2, $N_A$ the number with a bound, accessible and assessable artifact, and $N_M$ the number with no bound artifact. We report:

$$\text{Claim Support Rate}=\frac{N_2}{N}$$

$$\text{Assessment Coverage}=\frac{N_A}{N},\ \text{Missing-Binding Rate}=\frac{N_M}{N}$$

$$\text{Conditional Support Rate}=\frac{N_2}{N_A}$$

If $N_A = 0$, the conditional support rate is undefined and is reported as "not estimable," rather than zero.

The claim support rate is the primary measure. Its denominator is all evaluated material claims, so a system cannot improve its score by making evidence inaccessible or by leaving claims unbound. The conditional support rate shows how good the evidence is when it can be inspected, and it is always reported together with assessment coverage. The full distribution of scores (2, 1, 0, U, missing binding) is reported for every condition.

### 6.4 Evidence recoverability (RQ3)

The proposed recoverability evaluation distinguishes restoration from re-derivation. These tests are reported separately because retrieving a previously bound artifact and reproducing the process that selected it establish different properties. No completed evaluation under this protocol is reported here.

Restoration tests whether the exact artifacts bound to an output remain retrievable through their recorded identifiers, subject to the requester's authorization, without re-running retrieval, routing or generation. The evaluation requires artifact identifiers, recorded versions, content digests and the declared hashing method, including the bytes or canonical representation covered by each digest. A path succeeds only when every required artifact is recovered and matches its recorded version and digest. A matching identifier or a currently available replacement document is insufficient. Missing verification metadata makes the path unassessable, not successfully restored.

Re-derivation tests whether the recorded input and historical configuration reproduce the same bound artifact set. The replay configuration includes the corpus snapshot, index, embedding model, routing or selection method, policy settings and any other component affecting selection. Success requires equality of artifact identities, versions and digests. Missing historical inputs or configuration prevents a successful re-derivation claim. A relaxed top-k inclusion measure may be reported only as a separately defined secondary measure, with k fixed in advance; it does not count as exact re-derivation.

The same cohort of recorded paths is assessed immediately after generation and after a documented, controlled update. Post-update restoration checks continued access to the original artifacts. Post-update historical replay uses the original configuration or its preserved snapshot. Running the updated configuration instead measures sensitivity to system change and must be reported separately.

Let $P$ be the number of recorded evidence paths selected for evaluation. For each test condition:

$$\text{Restoration Rate}=\frac{\text{Paths restored with matching hash and version}}{P}$$

$$\text{Re-derivation Rate}=\frac{\text{Paths re-derived with matching artifact set}}{P}$$

Failed and unassessable paths remain in the denominator and are reported separately with reasons. Withdrawal, deletion and access restrictions are recorded explicitly; the test must not bypass authorization to improve recoverability.

Recoverability depends on retained artifacts, configuration and authorized access. It is not established by deterministic record verification. The cross-language verifier results in Section 8.3 concern agreement on record checks, not successful restoration or re-derivation.

### 6.5 Platform-domain classification and design alternatives (RQ1, RQ6)

**Scope and success criterion.** RQ1 is a closed-set top-1 classification task over the complete 15-class roster of the nage-8b router configuration evaluated on 21 April 2026. This roster is a subset of the wider platform's registered modules. The reference label is the domain associated with the input's validation dataset, not an independently annotated source attribution or a verified statement about which knowledge contributed to an answer.

In the retained routing_iq routine, each text is truncated or padded to 128 tokens and converted to embeddings using the frozen Qwen3-8B token-embedding layer. The domain router, STRIA, produces domain-weight vectors, called STEMMA, which the benchmark averages across token positions and routing blocks. A scored example succeeds when the largest component of the averaged vector corresponds to the dataset-domain label. The routine scores up to ten successfully processed non-empty examples per domain and skips processing failures. The retained summary reports 150 scored examples; attempted and skipped counts are not recorded.

This measure evaluates domain classification from routing weights. It does not evaluate a generated answer, customer-document selection, claim-to-source support, separate language-router performance or whether a delivered record identifies the module correctly. No verified historical split manifest or per-example evaluation ledger is available in the reviewed provenance. Exact training/validation membership, overlap with validation examples used during monitoring and cross-domain overlap therefore remain unverified (§7.2).

**Design alternatives.** RQ6 concerns experiment-specific outcomes in the native-model CRATON setting, rather than changes in the RQ1 classification score. A reliable-improvement claim would require repeated, matched evaluations of the baseline and alternative under the same task, scoring rule and evaluation population, with run counts, seeds and configuration differences reported. Improvement would be assessed against run-to-run variability using a prespecified analysis. The historical capacity evidence does not provide such a repeated-run comparison, so its results are descriptive and the reliability criterion could not be assessed. Corrected scores from later experiments retain their own configuration and scoring provenance; they are not presented as a rescoring of the original capacity-growth run.

### 6.6 Human evaluation rubric

The proposed human evaluation assigns one support score to each material claim identified under Section 6.2, considering the evidence bound to that claim and its recorded context. Where several artifacts are bound, they are assessed together for the claim-level score; artifact-level observations may be retained separately. Scores describe evidential support, not whether the model relied on the evidence during generation.

- **2 — Fully supported.** The available evidence supports the complete material claim, including its scope and qualifications, without an unresolved material contradiction.
- **1 — Partially supported.** The evidence supports part of the claim but does not adequately support a material element, scope restriction or qualification.
- **0 — Unsupported.** The assessable evidence provides no adequate support, is materially irrelevant or contradicts the claim. Contradiction is recorded separately; a score of 0 alone does not imply contradiction.

- **U — Unassessable.** Evidence is bound, but access restrictions, missing content or insufficient context prevent a defensible assessment. U is not a low support score.
- **Missing binding.** No evidence artifact is bound to the claim. This is recorded separately from U and the ordinal scores.

All evaluated claims remain in the denominator, including U and missing-binding cases. The complete outcome distribution is reported alongside the measures in Section 6.3.

Evaluators would receive the claim, its bound evidence and the context required for assessment, but not the system's γ scores, epistemic label or automated assessment scores. Before scoring begins, the evaluation must document the annotators' number, expertise and training, together with the size and selection procedure of the independently double-scored subset. At least two evaluators would score that subset independently. Agreement is calculated before adjudication; disagreements are resolved through documented discussion and, where needed, a third evaluator. Original and adjudicated scores are both preserved.

Automated evaluators may assist with triage or provide a separately reported analysis. Their model, version, prompt and relevant settings must be disclosed, and agreement with human judgments must be assessed on the double-scored subset [Li et al., 2024; Press et al., 2024]. Automated judgments alone do not constitute the human-rubric evaluation. No completed evaluation under this rubric is reported here.

### 6.7 Statistical reporting

Proportions are identified by their numerator, denominator, population and scoring rule. Where binomial intervals are reported, we use nominal 95% Wilson score intervals [Wilson, 1927], including for small samples and proportions near zero or one. For example, 137/150 (91.3%) has an interval of approximately 85.7–94.9%, while 0/15 has an upper bound of approximately 20.4%. These are descriptive summaries; they do not establish representative sampling, independence among related examples or certainty about historical execution provenance. AUC results, verifier-check counts and proposed evaluations are not assigned binomial success intervals. Small per-domain cells are not used for statistical comparisons.

For the proposed human evaluation, ordinal-score agreement would be reported using Krippendorff's alpha, with U and missing-binding cases excluded from the ordinal scale and their frequencies and agreement reported separately [Hayes and Krippendorff, 2007]. Agreement is measured before adjudication, and the size and selection of the double-scored subset must be declared. No completed rater-agreement study or composite score combining the measures is reported here.

### 6.8 Scope of the measures

These measures describe the behavior of the evaluated implementation under the stated corpus, benchmark, policies and threat scenarios. They are not global measures of truthfulness, safety or readiness for deployment. Authority fidelity would be measured partly through bounded response and label consistency; that evaluation is specified in Appendix A.2 and is proposed, so no part of authority fidelity is established by the results of this paper. The remaining part, namely who acts on an output and under what review, is a property of the deployment, and it is addressed through the documentation in Section 11 rather than through a benchmark.

## 7. Experimental Setup

This section documents the configurations, evaluation populations and procedures associated with the results reported in Sections 8 and 9. The historical routing benchmark, claim-assessment diagnostics, record-verification tests and native-model experiments use different configurations and are described separately. Table 7 identifies the retained configuration information for each experimental setting; additional distinctions are stated where experiments within a setting differ.

Each result is linked to the available version references, scoring method and supporting artifacts. Missing execution details are marked as not recorded or unverified rather than reconstructed from current defaults. Historical runs and later reproductions are identified separately. Proposed evaluations are described as protocols and are not presented as completed experiments.

### 7.1 Systems and configurations under evaluation

Table 7 separates the configurations associated with the historical routing benchmark, offline claim-assessment diagnostics, native-model experiments and synthetic-record verification. Each column identifies an experimental setting rather than a single deployment shared by every result. Where runs within a setting differ, their configurations and provenance are reported separately.

"Not applicable" means that a component is not used by the stated procedure. "Not recorded" identifies missing retained execution information. "Not verified" means that the available evidence does not establish the claimed historical identity. A retained source commit or artifact digest is not automatically the identity of the code or model executed during the original run.

**Table 7. Retained configuration information and unresolved execution details by experimental setting.**

| Component | RQ1 routing benchmark | RQ2 offline claim-assessment diagnostics | RQ6 native-model experiments | Synthetic verification fixtures |
|---|---|---|---|---|
| Source and execution references | Historical nm/fehm-v3.1; result reference 7eb41bb; benchmark source 41d3e96. Exact executing-tree digest not recorded. | Source reference d3aaca93. Separate 25 October detector reproduction: execution reference 02ed2d1c, result retained at 929a3e44. These do not pin the original execution tree. | Run-specific references; no common executing build. Original runs and later reproductions are distinguished in the technical annex [Ozturk, 2026]. | Executed revision for the reported 15 September fixture run is not established here. The later b34ced5 reference belongs to the separate 22 September verification study. |
| Generation model | Frozen Qwen3-8B base supplies embeddings; no answer is generated or scored. | Not applicable: existing claims and evidence strings are scored offline. | Approximately 252.7M-parameter Core and 642M-parameter configurations in the relevant model experiments. Packing-visibility tests do not themselves generate answers. | Not applicable. |
| Tokenizer | Qwen3-8B tokenizer; exact historical revision and file digests not recorded. | No generation tokenizer. Text processing is part of the versioned scoring methods. | craton-bpe-final.model is named in the records. The 25 October packing reproduction pins a particular tokenizer digest; it is not assigned retroactively to every historical run. | Not applicable to generation; Unicode handling remains part of verification. |

| Component | RQ1 routing benchmark | RQ2 offline claim-assessment diagnostics | RQ6 native-model experiments | Synthetic verification fixtures |
|---|---|---|---|---|
| Embedding or retrieval representation | Frozen Qwen3-8B token embeddings; exact base revision not recorded. | No embedding model used by the lexical scorer or mechanical detector. | BM25 selection does not use a neural retrieval encoder. Model token embeddings belong to the relevant checkpoint. | Not applicable. |
| Index | Not used by this routine. | Not applicable: evidence is supplied in each evaluation item. | BM25 over craton/data/store.db for the relevant retrieval and packing experiments; historical index-build identity is not fully pinned. | Not applicable. |
| Corpus and population | Fifteen domain-validation pools. Historical split membership and example identities remain unverified. Retained preparation recipe: seed 42, 80/10/10 split and per-domain prefix de-duplication. | Development: 105 items—29 supported, 64 contradicted, 12 silent. Holdout: 95 items—26 supported, 59 contradicted, 10 silent. | S3 contains 300 items, including 250 answerable items. Packing evaluates the answerable subset. The 15-case weight-path test is a separate population. | English synthetic sample and Unicode regression fixture; neither is a production-traffic sample. |
| Router or selector | STRIA v3.1; 2,400 steps; entropy weight 0.30; 75 dynamic samples/class. Reported checkpoint: stria_v3.pt, MD5 99d0d701fa176856ffa8b76658592a0f. | Not applicable: no retrieval or routing is performed by the offline scoring harness. | BM25 selector for the relevant context-based experiments. Weight-delta experiments use a separate inference path. | Not applicable. |
| Policy configuration | No policy decision evaluated. | No runtime policy decision evaluated. | No common runtime policy profile is established for these experiments; experimental acceptance gates are reported separately. | Runtime authorization is not evaluated. |
| Input or prompt construction | Dataset text/prompt field; no answer-generation prompt is evaluated. | Stored claim and evidence fields. Sentence-level detector input is identified in the 25 October reproduction. | Run-specific templates and packers. Legacy truncation and repaired token-budget packing must be distinguished. | Fixture construction is separate from model prompting. |

| Component | RQ1 routing benchmark | RQ2 offline claim-assessment diagnostics | RQ6 native-model experiments | Synthetic verification fixtures |
|---|---|---|---|---|
| Scoring method | Largest averaged routing weight compared with the dataset-domain label. | klm-support-lexical/0.1, klm-support-scoped/0.2 and klm-contradiction-mechanical/0.1. | Historical substring scoring, corrected word-bounded prefix scoring and packing visibility are separate measures. | Six record checks under the specified verifier profile; no generated-answer quality score. |
| Input or context budget | 128 token positions, truncated or padded. | No model context window; the supplied evidence string defines the scoring input. | Run-specific. Earlier legacy settings differ from repaired 8,000-token packing; the retained 4,000-token ledger is a separate run. | Not applicable. |
| Decoding and randomness | No answer decoding. Seed 42 belongs to the preparation recipe, not a verified record of every evaluation setting. | Deterministic scoring; no answer decoding. | Per-run generation settings must be reported separately. A common seed or batch-invariance guarantee is not established. | No model decoding. The sample's public test seed concerns fixture construction, not generation determinism. |
| Record schema evaluated | No emitted KLM record is evaluated. | No emitted KLM record is evaluated by these offline diagnostics. | These task and packing results do not establish a KLM record schema or conformance level. | Retained sample: klm-attestation/0.1. Other fixtures must be identified by their own declared profiles. |
| Verifier and canonicalization | Not used. | Not used. | Not used by the stated task and packing measures. | Retained sample declares sha256-canonical/2. Verifier revision and declared format must be linked to each fixture and run; later formats are not substituted. |
| Hardware and environment | RunPod A100 reported in the summary; runtime image and environment digest not recorded. | Historical local execution reported; original environment not pinned. | GPU training and local CPU/MPS evaluation differ by run. Hardware, dtype and software versions must not be pooled into one configuration. | Local execution reported; historical environment identity not established here. |

**Which system produced which result.** The RQ1 result is a historical platform-domain classification diagnostic on the frozen Qwen3-8B-based configuration; it is not a live API acceptance test. The RQ2 diagnostics score constructed claim–evidence inputs without generating answers. The native-model capacity, weight-path and packing experiments belong to CRATON and retain their own populations, scoring rules and execution limits. Synthetic verification tests examine record properties without a generation model. The production operator-view study in Section 8.3 is a separate case and is not merged with the synthetic fixtures.

### 7.2 Datasets and benchmarks

Table 8 identifies the populations used by the reported experiments and supplementary observations. Their units differ: scored inputs, claims, generated answers, evidence paths, records and stored chunks are not interchangeable. Some rows describe subsets or repeated assessments of the same underlying material; their counts must not be added as independent observations.

For RQ1, the retained preparation code specifies per-domain de-duplication using an MD5 digest of the first 500 characters, shuffling with seed 42 and an 80/10/10 train/validation/test split. This describes the preparation recipe, not verified membership of the historical splits. The author reported on 27 October 2026 that the training and validation files remain in a private archive; they were not independently inspected for this review, and no verified historical example manifest is available.

The planning, evaluation and data classes use keyword-filtered material from the same source dataset, yahma/alpaca-cleaned. Because de-duplication and splitting occur within each domain, an example could occur in multiple domain pools and in different partitions across domains. The training script also uses validation-pool examples for monitoring. Exact overlap with the scored benchmark examples remains unverified. These are limitations of the available provenance, not demonstrated explanations of the observed errors. The RQ1 result is therefore not established as performance on an untouched test set.

**Table 8. Evaluation populations and their scope.**

| Population | RQ or location | Size and unit | Construction and interpretation |
|---|---|---|---|
| Historical platform-domain routing benchmark | RQ1 | 150 scored inputs; 15 classes × 10 | Validation-file inputs; dataset-domain labels. The loader reads up to 100 non-empty texts per domain, and the scorer retains up to ten successfully processed examples. Historical membership and overlap remain unverified. |
| Claim-support development set | RQ2 | 105 claims: 29 supported, 64 contradicted, 12 silent | Constructed adversarial examples; labels follow construction rules. Not representative traffic or a completed human-rubric evaluation. |
| Contradiction-detector holdout | RQ2 | 95 claims: 26 supported, 59 contradicted, 10 silent | Constructed holdout, reported as separate from development tuning. Original execution and later reproduction are distinguished. |
| Live-answer diagnostic set | RQ2 | 236 answers: INVENTED 59, CLEAN 137, BORDERLINE 40 | Separately labelled live-generated answers evaluated on 30 September 2026. These labels are not interchangeable with the constructed-set labels; representative traffic sampling is not established. |
| Native-model S3 evaluation set | RQ6 | 300 items, including 250 answerable items | Constructed evaluation world. Answerable-item scores and packing measurements use the relevant 250-item subset. |

| Population | RQ or location | Size and unit | Construction and interpretation |
|---|---|---|---|
| Weight-path test | RQ6 | 15 held-out questions: mauritius 7; varlik 8 | Corrected QA test from two knowledge units. A health sweep selected lr 3e-4 and 300 steps before the corrected held-out test. The earlier selection-plan and instrument limitations remain explicit (§§7.4, 9.1); representative-use performance and repeated-seed reliability are not established. |
| Synthetic verification fixtures | Record checks in §8.3 | Two reported fixtures | English synthetic sample and Unicode regression fixture. Neither is a production-traffic sample; run-specific test scope must be stated. |
| Production disclosure record | Record checks in §8.3 | One operator view dated 30 August 2026 | Individual production record, separate from the synthetic fixtures and the September demonstration. |
| Live-demo observation | §8.3 | One identified record dated 28 September 2026 | Findings retained in a dated narrative report; original verifier output not preserved. |
| Source-hash backfill | Integrity checks in §8.3 | 7,230 stored chunks | Checks text held at backfill time; does not establish uninterrupted historical integrity or evidence restoration. |
| Final-input probe | §8.6 | One request | A retained probe observation; not a population-level estimate of evidence presence or reliance. |
| Controlled source-change observation | §8.6 | Five cases in the tested intervention arm | Partial result from an incomplete acceptance protocol; not the total population of every protocol arm. |
| Boundary-isolation evaluation | RQ4 | No completed evaluation population reported | Proposed protocol in Appendix A.1. |
| Bounded-response evaluation | RQ5 | No completed evaluation population reported | Proposed protocol in Appendix A.2. |

### 7.3 Evaluated domain roster

The 15 domains are the complete class set of the nage-8b router configuration evaluated on 21 April 2026. The roster was fixed through product development on 14 April 2026, expanding from four to eleven and then fifteen classes. It is a subset of the wider platform's registered module space. Undeployed registrations, separate 14B model-family configurations and marketplace modules were outside this evaluated router.

No predeclared statistical sampling or domain-selection procedure is recorded. The available provenance identifies training-data availability as a constraint on roster construction. Table 9 therefore describes each class's recorded role, not a separately documented reason for selecting it. Each class contributes ten scored examples.

The roster contains eight language-labelled classes, two code classes, two knowledge classes and three reasoning-related classes. These labels describe the router configuration; they do not establish language-specific generation quality, sector coverage or discrimination among modules outside the evaluated set.

**Table 9. Complete class roster of the historical nage-8b routing benchmark.**

| Router class | Scored examples | Recorded class role |
|---|---|---|
| fehm-tr | 10 | Turkish-language class |
| fehm-en | 10 | English-language class |
| ming-py | 10 | Python code |
| cortex-reasoning | 10 | Reasoning |
| fehm-ar | 10 | Arabic-language class |
| fehm-de | 10 | German-language class |
| fehm-fr | 10 | French-language class |
| fehm-es | 10 | Spanish-language class |
| ming-ts | 10 | TypeScript code |
| bilge-ml | 10 | Machine learning |
| bilge-data | 10 | Data science and SQL |
| fehm-ja | 10 | Japanese-language class |
| fehm-zh | 10 | Chinese-language class |
| cortex-planning | 10 | Planning |
| cortex-eval | 10 | Evaluation |

Historical SEDIM terminology groups five modules under CORTEX, whereas platform-family naming separates two Bilge classes from three Cortex classes. These naming conventions do not define different evaluated class sets. Throughout the paper, bilge-data denotes the Data science and SQL class.

### 7.4 Procedures

**RQ1.** Scoring was automatic in routing_iq at retained source reference 41d3e96. The loader retains the first 100 non-empty validation texts per domain; the scorer uses up to ten successfully processed examples per domain and compares the argmax of averaged routing weights with the dataset-domain index. No delivered record or generated answer is read. The retained summary does not preserve the exact invocation, executing-tree digest, individual predictions or skipped-example count.

**RQ2.** The dated method report describes development-set tuning followed by a holdout evaluation of the separate contradiction detector. Labels follow the adversarial construction rules; independent two-annotator agreement was not obtained. The original detector command and case-level output are not retained. The 25 October reproduction explicitly uses detect(claim, evidence_sentence) on the pinned 95-item holdout; whole-document input is a separate variant. The live-answer diagnostic of 30 September uses 236 separately labelled answers. Its contradiction-only D1 and contradiction-or-silence D2 variants are kept distinct. None of these procedures constitutes the proposed human-rubric evaluation.

**RQ3.** Historical reports describe Python–TypeScript agreement on synthetic fixtures, self-tests using deliberate corruptions, and a separate production operator-view check. The original draft reports three corruption classes for the 15 September fixture study, whereas the later bundle includes an additional removed-signature test. These are not pooled into a single four-class result on every fixture. Executed versions and original logs are not fully linked for the earlier fixture run. The dated 27 October author review reports a repeat of the production-view check associated with source b34ced5; it is separate from the original observation. No completed restoration or re-derivation evaluation is reported.

**RQ4 and RQ5.** Appendix A specifies proposed boundary-isolation and bounded-response procedures. No completed evaluation under those protocols is reported.

**RQ6.** The weight-path source record, clast-qa-lr-sweep-sealed.md §§9–10, reports a corrected set of 15 held-out questions from two knowledge units (mauritius 7; varlik 8). A health sweep selected lr 3e-4 and 300 training steps before the corrected held-out test. The record identifies a missing tie-break rule in the earlier selection plan and a corrected QA instrument. These qualifications are retained; no repeated-seed comparison is inferred.

The original capacity-growth study compares about 252.7M and 642M parameters under its historical training and substring-based scoring setup. Later baseline and Y1 word-bounded measurements are separate experiments, not a rescoring of the missing growth-run outputs. The corrected em_prefix_token method tests whether a gold/alias token sequence occurs contiguously in the answer prefix before the separator ⟵; it is not full-answer exact equality.

Packing tests count whether a gold-containing unit survives selection and packing under the stated matching rule. The historical substring rule, 2 October word-bounded rerun, separate retained 4,000-token ledger and 25 October 8,000-token reproduction retain distinct dates, budgets and artifact identities. A visibility flag does not establish semantic support or model reliance.

### 7.5 Predeclared criteria

Predeclaration is stated only where a dated criterion can be identified. The claim-support set design records K3 as discrimination between silent and contradicted examples with an AUC minimum of 0.65. This is not a per-claim support threshold and does not compare the supported class with the contradicted class.

The historical routing benchmark predates this paper's proposed protocol and is descriptive. The native growth and weight-path records contain their own dated gates; those are not a blanket pre-registration of every criterion in this paper. The weight-path correction preserves its revised task definition and the recorded selection-plan gap. The word-bounded scoring correction was introduced after inspecting outcomes and is explicitly post hoc. The 30 September live-answer detector test has its own dated pre-registration, including D2 as the primary variant. Later reproductions do not convert post hoc criteria into predeclared ones.

8. Results

Table 10 separates historical summaries, later reproductions, dated author observations and proposed evaluations. All findings are internal or author-reported; none is presented as independently replicated. Nominal 95% Wilson intervals are descriptive binomial summaries, not a remedy for constructed populations, source clustering, missing inputs or uncertain execution provenance. Verification counts describe checks or fixture classes, not independent sampled inferences. Proposed evaluations have no empirical result.

**Table 10. Results with population and provenance limits.**

| RQ / measure | Result | 95% interval | Population | Provenance and interpretation |
|---|---|---|---|---|

| RQ / measure | Result | 95% interval | Population | Provenance and interpretation |
|---|---|---|---|---|
| RQ1 domain classification | 137/150; 91.3% | 85.7–94.9% | Historical SB-2; 15 classes × 10 | Numerator recovered from rounded summary; no case-level prediction ledger. |
| RQ2 K3 ranking | AUC 0.008, lexical and scoped; minimum 0.65 | No AUC interval retained | 12 silent versus 64 contradicted within dev n=105 | Internal historical capture. Supported n=29 is not in this AUC. |
| RQ2 holdout detection | 52/59; 88.1% | 77.5–94.1% | 59 contradicted; holdout n=95 | Exact count from 25 Oct reproduction; reproduces historical rounded percentage. |
| RQ2 supported false positives | 3/26; 11.5% | 4.0–29.0% | 26 supported; same holdout | 25 Oct reproduction; related examples limit independence. |
| RQ2 silent false positives | 0/10; 0.0% | 0.0–27.8% | 10 silent; same holdout | 25 Oct reproduction; subgroup result, not overall false-positive rate. |
| RQ2 live-answer D1 | Detection 19/59; CLEAN flags 34/137 | Descriptive; source counts retained | 236 answers: 59 INVENTED, 137 CLEAN, 40 BORDERLINE | 30 Sep contradiction-only diagnostic; acceptance not met. |
| RQ2 live-answer D2 primary | Detection 33/59; CLEAN flags 46/137 | Descriptive; source counts retained | Same 236 answers | Contradiction-or-silence; primary registered variant; acceptance not met. |
| RQ2 human rubric | Proposed; no result | Not applicable | No completed human-rubric population reported | Automated diagnostics are not a human support-rate result. |
| RQ3 verifier agreement | Reported 6/6 check agreement per fixture | Not applicable | Two reported synthetic fixtures | Author-run historical report; exact run identity not fully linked. |
| RQ3 corruption self-tests | Earlier report: three classes; later bundle adds signature removal | Not applicable | Run-specific fixture scope | No combined 4/4-on-both-fixtures claim is made. |
| RQ3 production operator view | 1 PASS; 3 FAIL; 2 NOT MEASURED | Not applicable | One record dated 30 Aug | 22 Sep study; 27 Oct repeat reported by author; not accepted. |
| RQ3 source hashes | 0 mismatches / 7,230 chunks | Not applicable | Backfilled stored text | Database consistency; root unsigned and unannounced in historical record. |
| RQ3 restoration and re-derivation | Proposed; no result | Not applicable | No completed protocol evaluation reported | Record checks are not restoration or historical replay. |
| RQ4 isolation | Proposed; no result | Not applicable | Appendix A.1 | No completed live two-tenant acceptance result reported. |

| RQ / measure | Result | 95% interval | Population | Provenance and interpretation |
|---|---|---|---|---|
| RQ5 bounded response | Proposed; no result | Not applicable | Appendix A.2 | No bounding, abstention or label-consistency result reported. |
| RQ6 weight-path test | 0/15 | 0.0–20.4% | mauritius 7; varlik 8 | Historical corrected QA test; selected configuration, not repeated seeds. |
| RQ6 original capacity growth | Historical arm-A 8.0% → 8.8% | Not assessable | Original 252M → 642M setting | Legacy substring scorer; original outputs not retained for corrected rescoring. |
| RQ6 later corrected measurements | Baseline 1/250; Y1 arm A 3/250 | Not used for a paired growth inference | Later experiments; 250 answerable items | Different configurations; not corrected outcomes of the original growth run. |
| RQ6 legacy packing loss | 118/221; 53.4% | 46.8–59.9% | Selected-answer subset of 250 | Historical normalized-substring visibility. |
| RQ6 repaired packing | 221/221; 100% | 98.3–100% | 8,000-token selected-answer subset | Historical aggregate; original per-item ledger missing; separate 25 Oct reproduction retained. |
| RQ6 word-bounded packing rerun | Lost 130/219; repaired 219/219 | Loss 52.7–65.7%; retention 98.3–100% | 2 Oct; selected-answer subset of 250 | Separate matching rule and run; not answer accuracy. |
| Supplementary source change | 5/5 new value; 0/5 old value | 56.6–100% for 5/5 | Tested intervention arm only | 20 Sep partial protocol; overall acceptance incomplete. |
| Supplementary input probe | One target span: tokens 945–950 / 1,264 | Not applicable | One retained live request | Input-presence observation; not a reliance measurement. |

### 8.1 RQ1: Platform-domain classification

The retained 21 April 2026 SB-2 summary reports accuracy 0.9133 over 150 scored inputs from 15 classes. With four-decimal rounding, 137 is the compatible integer numerator: 137/150 = 91.3% (nominal 95% Wilson interval 85.7–94.9%). The reference labels are dataset domains, not independent source-attribution judgments.

Table 11 separates the three retained weak-domain rates from recovered counts and the inferred aggregate of the other twelve classes. Their individual predictions and confusion types are not retained. The errors are argmax/domain mismatches under Section 6.5; the test reads neither a delivered record nor a generated answer. Historical split membership, training-monitoring overlap and cross-domain overlap remain unverified. This result must not be substituted for the later RoutingIQ score of 94.8 or the composite SB-2 score, which are different reported quantities.

**Table 11. Historical SB-2 platform-domain routing: retained rates and recovered counts.**

| Domain or group | Correct / scored | Error count and evidence status |
|---|---|---|
| bilge-data | 7/10 (recovered) | 3 argmax mismatches; summary accuracy 0.7. |
| cortex-planning | 5/10 (recovered) | 5 argmax mismatches; summary accuracy 0.5. |

| Domain or group | Correct / scored | Error count and evidence status |
|---|---|---|
| cortex-eval | 5/10 (recovered) | 5 argmax mismatches; summary accuracy 0.5. |
| Other 12 domains, aggregate | 120/120 (inferred) | Remainder from total and three listed domains; individual domain results not retained in summary. |
| Total | 137/150 (recovered) | 13 argmax mismatches; per-example predictions and detailed confusion types not retained. |

### 8.2 RQ2: Claim-assessment diagnostics

The evaluation separates overlap-based support scoring, a mechanical contradiction detector and the proposed human-rubric evaluation. The historical development set has 105 constructed claims: 29 supported, 64 contradicted and 12 silent.

The lexical variant assigns means of 0.890 to contradicted and 0.780 to supported claims. Separately, K3 compares the 12 silent with the 64 contradicted examples; supported claims are excluded. The retained harness computes the fraction of silent/contradicted score pairs in which the silent score is greater, counting a tie as one half. Both lexical and scoped variants report AUC 0.008 against the registered minimum of 0.65. The class means and the K3 AUC are distinct observations; neither is a calibrated production support decision.

The deterministic detector checks numbers, dates, polarity and entities behind a topic-overlap gate. The historical report describes development tuning and one 95-item holdout evaluation, with rounded rates of 88.1% detection, 11.5% false positives on supported claims and 0.0% on silent claims. The original command and case-level detector output are not retained. A separate 25 October reproduction on the pinned holdout with sentence-level evidence gives 52/59, 3/26 and 0/10. These exact pairs are reproduction counts, not recovered original outputs. Whole-document input in that reproduction gives different results and is not substituted. Nominal Wilson intervals and class denominators appear in Table 10; source-related examples limit the independence assumption.

The separate 30 September live-answer diagnostic contains 236 labelled answers: INVENTED 59, CLEAN 137 and BORDERLINE 40. D1, contradiction only, detects 19/59 INVENTED answers and flags 34/137 CLEAN answers. D2, contradiction or silence and the primary registered variant, detects 33/59 and flags 46/137. Neither meets its acceptance criterion. The labels and procedure differ from the constructed holdout; these results do not establish representative-traffic performance. In particular, the constructed-set detection rate does not justify live $\gamma$ integration.

Labels in the constructed sets follow transformations, not an independently double-annotated human support study. No completed evaluation under Section 6.6 is reported, so no human-rubric support rate, coverage distribution or rater-agreement result is claimed.

### 8.3 RQ3: Evidence recoverability and record verification

**Record verification.** The historical report describes identical Python–TypeScript verdicts on six checks for each of two synthetic fixtures in a local study dated 15 September 2026. Both implementations were written within the authors' organization. This is reported fixture-level agreement, not independent replication or evidence of operating-level conformance. The original executing version and complete run artifact are not fully linked in the reviewed evidence. The earlier three-class corruption report and the later self-test's removed-signature case retain separate scopes; they are not combined into a four-class result on both fixtures.

**Production disclosure record.** The dated 27 October author review identifies the separate 22 September study with verifier source b34ced518a58fa44bb10d88b7b62189bc61a33ef and one real operator-view record dated 30 August 2026, inference aud_EUF9qAT0GXu2LC5x6ljvug. Its retained file digest is SHA-256 c4b955a5ca7f81be85d51c005f3371b12fb7ad9c3146240e3e92f6a6b7fefdb3.

The review reports an unchanged 27 October repeat. The record is not accepted: authenticity PASS; content integrity, honest nulls and composition manifest FAIL; γ invariant and claim-support recomputation NOT MEASURED. This is parity of adverse or unavailable verdicts, not successful acceptance.

The author review counts four unreasoned nulls in the jurisdiction and effective-date fields of two internal policies. The γ values are present: declared confidence 0.0, grounded_confidence.value 0.0 and gap 0.0. The historical verifier expects scalar grounded confidence and cannot extract the object-valued representation. NOT MEASURED is therefore a profile-extraction limit, not proof that γ is absent or arithmetically inconsistent. This zero-valued record would not be a discriminating arithmetic test even if extracted.

**Separate live-demo observation.** A 28 September narrative report, BULGU-CANLI-DEMO-OLCUMLERI-2026-09-28.md B4–B5 at 6b147a8, identifies record aud_22mPga5fLZSGM1FCMxbEaQ and reports nine unreasoned nulls and a γ extraction failure. It is not the August production record. The original verifier output and exact executed revision are not preserved in that report. The author confirms that a receipt-body copy exists privately and is unpublished; it has not been supplied as a public reproduction bundle. The earlier count of six unreasoned nulls has no identified source and is not retained as a finding.

**Source-hash consistency.** The retained 14 September SQL transcript reports 7,230 chunks backfilled, 7,230 marked backfill and zero mismatches against PostgreSQL pgcrypto recomputation. This is a consistency check using that digest implementation, not a three-implementation backfill test. The root recorded for 7,230 leaves is sha256:431bcbe82b8a466ead9951e9bc5b75a17a6e48b70ce29346b664cb3f1de4e3af. The original leaf manifest was not exported. A separately dated 25 October database re-fold is reported to reproduce the root; it is not the original manifest. No signed, externally announced commitment or published inclusion proof is established by the reviewed record.

**Restoration and re-derivation.** No completed evaluation under Section 6.4 is reported. Verifier agreement and hash consistency do not establish that bound evidence can be restored after an update or that historical retrieval can be replayed. Serving-path removal and durable retention also require separate treatment, as discussed in Section 11.5.

### 8.4 RQ4: Knowledge-boundary isolation

Appendix A.1 specifies a proposed evaluation covering cross-boundary scenarios and distinct violation classes. No completed evaluation under that protocol, including live two-tenant isolation acceptance, is reported. Policy or scope metadata and a source-change result do not substitute for this test. No isolation rate or general security claim is inferred.

### 8.5 RQ5: Bounded response

Appendix A.2 specifies a proposed evaluation of appropriate bounding, inappropriate abstention and label consistency. No completed evaluation under that protocol is reported. An epistemic label or a few examples of refusal do not establish calibrated bounded-response behavior or authority fidelity.

### 8.6 Evidence-path observations

**Final-input probe.** A retained minimal export dated 25 October preserves metadata for one request created on 20 September 2026. It reports a target span at token positions 945–950, 1,264 tokens before and after truncation and truncated=false. This is evidence of the recorded probe's observation on that request, not proof that every input is instrumented correctly or that the model used the span. The export omits the payload and signed receipt body; build_marker=dev and git_sha=null do not pin an immutable measured deployment, and the request-linked tokenizer files are not pinned.

**Controlled source change.** The retained 20 September run records five responses using the modified value and none using the old value in intervention arm B (5/5; nominal Wilson interval 56.6–100%). Arm A records four successful client requests out of five. The cross-tenant arm D was not run, and the overall sealed acceptance remains incomplete with exit code 3. This is bounded behavioral sensitivity to a source change, not an accepted isolation test or general proof of causal attribution. The local runner reports four dirty-tree status entries; no run-time diff or snapshot was retained, and the measured service's deployment identity was not captured. These provenance limits remain attached to the result.

## 9. Negative Results and Design Alternatives

This section reports experiment-specific failures and measurement defects. They are not four outcomes from one common benchmark: the weight-path task, capacity-growth study, claim-assessment diagnostics and packing checks use different populations and measures. Table 12 separates observations from proposed design responses and remaining tests. A failed implementation or scoring instrument is not a general impossibility result.

### 9.1 Weight-path configuration for route-to-one behavior

Route-to-one is an experimental inference path that selects one knowledge unit for a query and applies its associated weight delta during answer generation. It is separate from RQ1's platform-domain classification. The historical QA-delta study first used an instrument later found defective, then reports a corrected held-out set of 15 questions from two units: mauritius (7) and varlik (8). The selected configuration retained lr 3e-4 and 300 training steps after a health sweep; it was not selected by searching over the corrected held-out outcomes.

The corrected record reports zero clean answers and zero matching answers in all 15 cases (0/15; nominal Wilson interval 0.0–20.4%). The recorded Q1′ gate requires at least 9/15 clean answers and at most 3/15 source-echo or question-echo cases; it was not met. Source-specific terms in outputs are not counted as successful question answering. This finding concerns that recipe, two sources and the corrected task. It does not show that no weight-delta method can work, identify a unique cause of failure or establish performance on representative customer data. The record's small training set and selection-plan limitations remain relevant; no repeated-seed reliability result is claimed.

### 9.2 Capacity expansion

The original CRATON growth experiment expanded a roughly 252.7M-parameter Core configuration to about 642M parameters and historically reported arm-A scores of 8.0% and 8.8%. Arm A is the condition in which the selector supplies retrieved units to the generation model. These percentages used a normalized-substring answer-prefix scorer, which could accept short numeric gold strings embedded in longer numbers or placeholder indices. A later correction identifies this measurement defect. The original growth-run outputs are not retained for corrected rescoring, so the 0.8 percentage-point difference is not presented as validated capacity evidence.

Separate later experiments provide word-bounded prefix scores, including a 252M baseline at 1/250 and Y1 arm A at 3/250. They have their own training, packing and execution histories and are not corrected measurements of the original paired growth run. The historical growth record also notes a training-data defect: only 44% of training examples contained the answer in context. Capacity, training quality, scoring and evidence delivery are therefore not isolated by one comparison.

The available evidence does not establish a reliable capacity benefit under a repeated, matched evaluation. It also does not justify the claim that model capacity was definitively excluded as a bottleneck. Packing defects are demonstrated separately in Section 9.4; they do not supply the missing controlled capacity comparison.

### 9.3 Lexical claim support

Two observations must remain distinct. The lexical variant assigns a higher mean score to constructed contradicted claims than to supported claims (0.890 versus 0.780). K3, however, compares silent with contradicted examples: lexical and scoped variants each report AUC 0.008 for that comparison against the registered minimum of 0.65. Neither AUC is a supported-versus-contradicted result.

These diagnostics show why shared vocabulary alone is insufficient as a support decision rule on this construction. A changed number, date, polarity or entity can preserve vocabulary while reversing a statement's meaning. The design response is to keep support and contradiction distinct, preserve each record's method identity and retain the adverse examples as regression targets. Creating a separate detector is not the same as demonstrating that it transfers to live answers or integrating it into a deployed score. The 30 September D1 and D2 failures are therefore reported alongside the constructed-set result.

**Dated implementation report.** The implementing author's 27 October 2026 review reports that klm-support-lexical/0.1 remains a supplementary field within the grounded block and contributes to context_provenance, but does not enter grounded_confidence.value or the epistemic label. This supersedes the earlier description that the lexical figure was a numerical term in the grounded calculation. The same review reports that the contradiction detector was not connected to live $\gamma$. This is a dated author account, not an independently reproduced production finding: the underlying query/export and immutable record-to-platform revision linkage were not retained. Its aggregate record count is not introduced as a new measured result here.

The recorded status measured describes a mechanical computation; it does not make lexical overlap a validated entailment measure. A high support share and the absence of a contradiction flag are insufficient to establish support. The dated account does not establish an integration or withdrawal date, and no later runtime behavior is inferred from it.

### 9.4 Evidence lost before the model

In the native S3 answerable subset of 250 items, the historical normalized-substring visibility check selected a gold-containing unit for 221/250. Legacy packing retained such text for 103 of those cases, leaving 118/221 selected cases without it (53.4%; nominal Wilson interval 46.8–59.9%). At an 8,000-token budget the historical filtered stdout reports repaired packing retention of 221/221 (98.3–100%). These are conditional text-presence counts, not semantic relevance or answer-quality judgments.

The substring rule can count a short gold such as 6 inside 16 or 600. A separate 2 October word-bounded run reports selection for 219/250, legacy retention for 89/250, loss of 130/219 (59.4%; 52.7–65.7%) and repaired retention of 219/219. The qualitative packing loss remains under this stricter rule; the two sets of counts are not pooled.

Artifact scope matters. The original 8,000-token per-item gate8k.json was not retained; its filtered aggregate survives in a session transcript. The retained historical per-item ledger is a 4,000-token run with repaired retention 218/221, not the original 8,000-token result. The separate 25 October 8,000-token reproduction supplies a ledger and matches the historical substring aggregates on the pinned input artifacts, but uses later code. It does not recover the original working tree or original case-level ledger.

These checks demonstrate that selection and final-input presence can diverge in the tested pipeline. They do not show that a model understood the visible passage or that packing repair improved its answers. A matched evidence-use evaluation and any further reading-loss study remain separate requirements; no missing result is replaced by visibility alone.

### 9.5 What the negative results changed

The findings motivate separating measurements that can fail independently: route classification, support scoring, contradiction detection, packing visibility, answer behavior and record verification. The responses in Table 12 are requirements or scoped engineering changes, not evidence that the live system has closed every gap.

**Table 12. Findings, bounded implications and remaining evaluation.**

| Finding | What is established | Design response | Evidence still required |
|---|---|---|---|
| Weight-path 0/15 | The selected corrected QA-delta recipe failed on the reported two-source test. | Preserve task, source and recipe identity; do not equate loading or topic transfer with useful answering. | Larger, independently held-out sources and questions; repeated runs; separate removal and isolation tests. |
| Historical growth 8.0% → 8.8% | Legacy scoring and training defects limit the capacity comparison; original outputs cannot be corrected. | Separate growth results from later word-bounded experiments; audit scoring and training inputs. | Matched repeated comparisons with retained outputs, unchanged scoring and controlled training conditions. |
| Overlap diagnostics and failed transfer | Class means and K3 expose different overlap limits; the live-answer detector did not meet its gates. | Version support methods; record contradiction separately; keep adverse cases as regression targets. | Representative-data calibration and integration acceptance before a production support claim. |
| Packing loss | Both substring and word-bounded checks show selected evidence lost before the model in the tested setup. | Record selection, packing and final-input presence separately. | Matched evidence-use and answer-quality comparisons after repair; preserved execution identities. |

## 10. Discussion

### 10.1 What the results show about evidence visibility

The results support a bounded claim: inspectability depends on several distinct record and execution properties, and a successful intermediate metric need not establish the next property. The SB-2 summary supports historical domain-label classification under the stated closed-set rule; it does not evaluate a delivered record. The packing checks show losses between selection and model input. The claim-assessment diagnostics expose limits of lexical scoring and detector transfer. Record checks permit a different kind of inspection and include adverse production-view verdicts. No completed restoration, boundary-isolation or bounded-response evaluation is reported. These observations justify separate measurements, not a composite judgment that the whole service is evidence-visible or ready for deployment.

### 10.2 Why the negative results matter

The failures and measurement corrections are informative because they expose different limits of the original interpretations. Shared vocabulary was not sufficient evidence of support; retrieval did not ensure final-input presence; and the growth study did not isolate a reliable capacity effect. Historical experiments were not all executed under this paper's later protocol, so they must not be presented as a unified pre-registered validation of it. Their value is diagnostic: they motivate tests that can distinguish pipeline stages and retain adverse outcomes without retrospectively changing their execution history.

This has a governance implication beyond this system. Capability claims for AI systems are commonly supported by positive results alone. A procurement officer, auditor or regulator reading such claims cannot tell whether failures were absent or simply unreported. Requiring negative results, with their populations and design consequences, is a low-cost way to make capability claims more informative. It is a practice that standards and procurement requirements could adopt directly.

### 10.3 What existing evaluation practice would and would not capture

Table 13 compares the information required to identify the reported failure types. It is not a head-to-head evaluation of ALCE, RAGAS or any other implementation. An evaluation can detect a failure only when its inputs, reference material and scoring procedure expose that failure; extensions to an existing method may add such information.

**Table 13. Failure types and the evaluation designed to detect them.**

| Failure type | Answer and retrieval measures | Citation or faithfulness assessment | Evidence in this paper or proposed test |
|---|---|---|---|
| Selected evidence lost during packing | Answer accuracy may reveal an error without localizing it; retrieval measured before packing can still report a hit. | May expose the loss if the actual final input is supplied and evaluated. | Observed through scoped final-input visibility checks; not an answer-quality result. |
| A contradicted claim receives a high overlap score | A correct-answer reference may reveal the wrong claim; retrieval similarity alone does not determine support. | Human or automated semantic checking can test support, subject to evaluator reliability. | Constructed scoring diagnostics and a separate live-answer detector test; the human rubric remains proposed. |
| A cited artifact cannot be restored after change | Ordinary answer or retrieval scores need not test preservation of the historical artifact. | A current citation check may detect a broken link but does not necessarily test exact historical recovery. | Exact restoration and historical re-derivation under Section 6.4; proposed. |
| The record names the wrong boundary | Requires ground truth for the governing boundary and the emitted record. | Not established by citation support alone. | Violation class V4 in Appendix A.1; proposed. RQ1 does not inspect the delivered record. |
| Inappropriate abstention | Requires answerable controls and an explicit abstention criterion. | Evidence support can inform the control judgment but does not alone score whether an answer was withheld. | Control cases and label/behavior assessment in Appendix A.2; proposed. |

The comparison motivates record-level evaluation without claiming that existing approaches are incapable of being extended. Entries for proposed tests describe intended measurement scope, not observed detection performance.

### 10.4 Implications for governance practice

For deployers, the practical lesson is to ask for records, not descriptions. A procurement or assurance process can ask a vendor for a sample of real output records, have them checked with an independent verifier, and request restoration of the evidence for an output produced before the most recent system update. These are concrete tests that a buyer can run, and they distinguish demonstrated properties from documented ones.

For auditors and oversight bodies, the key test is restoration after change. Most audits happen after a system has been updated. A system whose evidence cannot be restored after an update cannot support the after-the-fact review that record-keeping obligations assume [European Parliament and Council, 2024].

For individual users, repeatable checks on a specified record can support examination of the answer they received without regenerating it. That assurance depends on fixed verification inputs and adequate disclosure; it is not a promise that every repeated question produces the same answer or that every historical source remains accessible. Artifact restoration and successful verification remain separate requirements.

For standard setters, the results locate a gap that current frameworks leave open. The NIST AI RMF, ISO/IEC 42001 and the EU AI Act define what an organization must be able to show. An output-level record specification is one way to make those obligations testable for individual outputs. Whether the KLM, or another specification, should fill that role is a question for independent implementers and standards bodies, not for its steward alone.

### 10.5 Implications for the KLM

The specification reference used here is KLM Specification v0.3, distributed under repository tag v0.3.0. Reasoned nulls, invalid-versus-absent handling, exact formula identifiers, separate contradiction signals and version-aware conformance claims are requirements in that reference, not newly demonstrated implementation capabilities. Earlier records retain the rules of their declared versions.

The packing observations motivate explicit final-input presence information with its own method and scope. The proposed human evaluation and user-facing interpretation studies remain distinct from those record requirements. Plain-language label explanations and authorized access to supporting evidence are design recommendations here, not a claim that this paper has established a universal display rule or measured its effects. Neither specification adoption nor a new formula identifier closes an untested production acceptance gap.

## 11. Limitations, Governance, and Responsible Interpretation

This section states what the results do not show, which governance risks evidence-visible systems create, and what remains the responsibility of those who deploy them. Its claims carry the same weight as the results in Section 8. The section applies the paper's own standard to the paper: a property counts as established only under a stated test condition, success criterion and failure boundary.

### 11.1 What evidence visibility does not establish

**Evidence visibility is not factual correctness.** Bound evidence may be outdated, incomplete, non-authoritative, misread or insufficient for the conclusion. Even a successfully restored artifact would establish availability of the recorded basis, not that it was the best evidence or that the answer is correct.

**Routing weights and claim–source links are not a complete account of model-internal reasoning.** They describe recorded selection or asserted evidence relations under stated methods. The 5/5 source-change observation is limited behavioral sensitivity in one tested arm, not general proof that every claim relied on the displayed evidence. Customer-document context, model-side modules and customer-specific learned weights retain separate acceptance requirements.

**Record integrity is not record completeness.** A verified signature authenticates the signed object under the trusted key, and a matching recomputed digest supports integrity of the covered content. Neither establishes that every request was recorded, that the recorded account is complete or that no alternative record exists. A signature on a claimed digest without the covered body does not by itself validate that body. Omission evidence requires additional mechanisms such as externally witnessed signed checkpoints and inclusion/consistency proofs. No such completed audit-trail claim is established by the unsigned historical Merkle root reported here.

### 11.2 Limits of the evaluation

The findings are internal and author-reported. Historical evaluations were designed or run within the developing organization, and none is presented as independently replicated. Separately written Python and TypeScript implementations are not organizationally independent. A dated source review or reproduction is also distinct from the original run and is labelled accordingly [Raji et al., 2020].

RQ1 contains 150 scored examples over a product-defined 15-class configuration, with dataset-derived labels and an inferred numerator. It does not establish source attribution, delivered-record accuracy or performance over the wider platform module space. The author reports historical data in a private archive, but the reviewed record does not verify historical membership, source revisions, validation-monitoring overlap or cross-domain overlap. Nominal Wilson intervals do not resolve those provenance uncertainties or establish representative-traffic performance.

The constructed claim sets have small subgroups and related examples. Construction-based labels do not substitute for independent human annotation. The 236-answer diagnostic is a separate labelled collection, not established as a representative traffic sample. No human-rubric support-rate or rater-agreement result is reported.

The capacity evidence does not provide a repeated matched comparison under a corrected scorer, and later word-bounded outcomes are not retrospective corrections of the missing original outputs. The weight-path result concerns a selected configuration and two knowledge sources; its 0/15 interval is a nominal small-sample summary, not a universal limit. Packing visibility is a text-matching diagnostic rather than proof of semantic support or evidence use.

The router roster contains language-labelled classes for Turkish, English, Arabic, German, French, Spanish, Japanese and Chinese, plus Python and TypeScript code classes and knowledge/reasoning classes. These class names do not establish independently measured generation quality in each language. Valen is a Turkish fictional source, and the constructed claim diagnostics include source-dependent language phenomena; no balanced cross-language evaluation is claimed.

Restoration, re-derivation, boundary isolation and bounded response remain proposed evaluations. Partial source-change and probe observations do not replace them. The public audit package is incomplete, limiting independent inspection and reproduction as stated in Section 11.9.

### 11.3 Limits of the security and boundary results

No completed boundary-isolation result under Appendix A.1 is reported. The presence of scope metadata, policy references or signed records does not establish enforcement against cross-tenant access, indirect injection, data exfiltration or other untested attacks. Any later result must be scoped to its controls, adversary model, input population and violation classes. In particular, controlling retrieved context does not remove knowledge already encoded in the base model's parameters. Customer-specific learned-weight realization and its isolation/removal acceptance remain separate, unestablished claims.

### 11.4 Limits of bounded response and epistemic labels

Bounded-response behavior depends on task framing, evidence, policy and model configuration. The proposed protocol measures both unsupported answering and excessive abstention, but no completed result is reported. Its eventual trade-offs would remain deployment-specific; an epistemic label is not an accepted decision boundary.

Epistemic labels compress a detailed assessment and may be over-trusted. A label should be accompanied by a plain-language account of its method and limits and an authorized route to available evidence; these are recommendations, not measured improvements in user decisions. Whether labels improve reliance for experts or individual users is an empirical question not tested here [Parasuraman and Riley, 1997].

The 27 October author account reports a lexical diagnostic inside the grounded block as supplementary information, not a term in grounded_confidence.value or the epistemic-label calculation. Its measured status describes a mechanical computation rather than entailment. The reported absence of live contradiction integration and the provenance limits of that status review remain explicit in Section 9.3. No later operational acceptance is inferred.

### 11.5 Governance risks of evidence-visible systems

Evidence-visible design reduces some governance risks and introduces others. We name four.

**Audit-washing.** A system that produces records, labels and signatures can look accountable without being so. Inadequate or self-conducted audits can give false assurance and lend legitimacy to practices they do not actually examine [Goodman and Tréhu, 2022]. A KLM record makes inspection possible. It does not show that anyone inspected, or that the inspection was adequate. Organizations should treat the existence of records as a precondition for accountability, not as evidence of it.

**Displaced accountability.** The system supports review. It does not replace review. For rights-, safety-, mission- or legally consequential decisions, authorized people remain responsible for validating evidence, assessing context and exercising judgment. Evidence-visible output can make that responsibility easier to carry out. It cannot transfer the responsibility to an automated system, and a well-formed record must not be used to suggest that it has.

**Tension between retention and erasure.** Recoverability can depend on retaining evidence that may contain personal, confidential or withdrawn material. The legal considerations discussed here are not resolved by a record format [European Parliament and Council, 2016; 2024]. In the reviewed implementation evidence, serving-path removal does not demonstrate complete deletion across stores, replicas, derived artifacts or object storage. The treatment of a previously signed record after evidence withdrawal, including any marker, retained hash or restricted restoration response, is not established by a completed acceptance test. It must be disclosed as an implementation limitation rather than filled with an assumed behavior. The proposed restoration protocol must respect authorization and report withdrawn, deleted or inaccessible artifacts without counting them as successfully restored.

**Concentration in the specification.** KLM is openly licensed, while the documented v0.x amendment process retains decision authority with its maintainer. Public proposals, validators and versioned rules support scrutiny but do not establish shared stewardship or independent evaluation. The reviewed reference implementation and verifier evidence remain associated with the specification's developing organization. Independent implementations, external evaluation and broader participation in decisions are future work, not properties demonstrated by this paper.

### 11.6 Responsibilities that remain with deployers

Authority fidelity is not established by the results of this paper; the bounded-response evaluation that would measure part of it is specified in Appendix A.2 (Section 6.8) and is proposed. The rest depends on how a system is deployed, who acts on its output, and under what review. Table 14 lists the responsibilities that no output specification or implementation can discharge, together with the function of the NIST AI Risk Management Framework under which each would ordinarily fall [NIST, 2023].

**Table 14. Deployment responsibilities outside the scope of the KLM.**

| Responsibility | What the deployer must decide or document | NIST AI RMF function |
|---|---|---|
| Intended and prohibited uses | Which decisions the system may inform, and which it must not | Map |
| Human review points | Which outputs require review before use, by whom, with what authority | Govern |
| Escalation | What happens to outputs labeled UNCERTAIN, CONTESTED or OUTDATED | Manage |
| Thresholds | The balance between appropriate bounding and inappropriate abstention for each use | Measure |
| User-facing display | How labels and evidence are shown to users, including individual users | Map, Manage |
| Retention and erasure | How long records and bound evidence are kept, and how withdrawal is handled | Govern |
| Incident handling | How a wrong or harmful output is investigated using its record | Manage |
| Independent review | Whether, by whom and how often records are examined by a party outside the deploying team | Govern, Measure |

A deployer that relies on KLM-conformant output without addressing these responsibilities has adopted a record format, not a governance arrangement.

### 11.7 Conflicts of interest and their mitigation

Omer Asim Ozturk is the founder of Nage AI, which develops SEDIM and stewards KLM.

Cihan Aydiner serves as a strategic advisor and AI red-teaming architect to Nage AI.

The methodological safeguards reported here are the retention of adverse outcomes, separation of historical and proposed evaluations, explicit population and provenance limits, versioned scoring and record rules, and disclosure of incomplete artifact access. Predeclaration is asserted only for identifiable dated criteria, not for every historical experiment. The human protocol keeps unavailable evidence in its intended denominators, but that protocol has not produced a completed result. These safeguards make limitations visible; they do not remove the conflict. Independent replication and review by parties without a stake in the outcome remain necessary next steps.

The corresponding author information: **me@omerasim.com**

### 11.8 Responsible interpretation

Readers who cite or rely on these results should observe four constraints:

Cite each result with its numerator, denominator, interval, population and evaluation label. For example: “137 of 150 platform-domain classifications in the historical SB-2 benchmark (internal, author-reported; 95% interval roughly 86% to 95%),” not “91% source-attribution accuracy.”

Attribute each result to its tested implementation setting and configuration, not to KLM as a general capability. Distinguish the historical platform-domain router, offline claim diagnostics, synthetic and production record checks, and CRATON experiments. The specification defines record requirements; these results concern particular implementations and tests.

Do not extend the results beyond the tested domains, corpora, languages, threat scenarios and configurations.

Do not treat KLM conformance, a signed record or an epistemic label as evidence of correctness, security or regulatory compliance.

### 11.9 Reproducibility and artifact availability

Availability is stated at the artifact level. A local file, private repository, package prepared for deposit or reserved DOI does not establish public release. No on-request access arrangement is implied unless a channel and conditions have been confirmed. The following status describes the evidence reviewed for this revision; it does not promise a later release.

**Public specification and validator.** KLM Specification v0.3 and its conformance material are available in omasim/KLM under tag v0.3.0. The reference validator distribution is klm-conformance 0.3.0. The specification document version, repository/package release, wire-schema identifiers and formula identifiers are separate. The historical Python/TypeScript verification bundle is a distinct artifact; its exact evaluated release must not be assumed included in the PyPI package or public specification tag.

**Historical verifier bundle and records.** Local copies of verify.py, verify.ts, the sample and parity test are retained, including a prepared publication branch. A public location for the exact historical bundle evaluated here has not been confirmed in this review. The test-keyed sample must not be confused with the privately retained production disclosure. The August operator view has digest c4b955a5ca7f81be85d51c005f3371b12fb7ad9c3146240e3e92f6a6b7fefdb3. The September demonstration body is also reported retained privately and unpublished. No public receipt-and-verifier reproduction bundle is claimed.

**RQ1 summary and provenance.** The retained results/sedim_bench_v3_1.json is dated 21 April 2026, with SHA-256 b9e3a6840880c093211c60270213c525f85e86d32c76675e25f770efe0b92773, source result commit 7eb41bb524d7a4ec63d67995ef3df0bb6e49c311 and benchmark reference 41d3e969b7f5a09e7eed3809d2609605fdba9e26. The three 27 October provenance records cover the recovered 137/150 count, data/split availability and roster selection. A private evidence package has been prepared, but no public release of these four files is established here. The author reports historical training and validation files in a private archive; they have not been inspected to verify exact historical membership. A verified split manifest, input hashes, source-dataset revisions and per-example prediction ledger are not supplied. A later training run would be a new experiment, not reproduction of the historical inputs by assertion.

**RQ2 evidence.** Retained private references include claim-support-eval-sonuc.md, the constructed input files at nage-platform d3aaca9328fe7ad2b444d9cf9c4228404d4cdd45 and the 25 October holdout detector reproduction at result commit 929a3e44a7fca05fe2b9dcab978147f6b5f2488e. The 30 September live-answer diagnostic has a separate result report and JSON. Original detector command/output retention is limited as stated in Section 8.2. The 27 October production-status account has no retained underlying query/export or immutable record-to-platform link. No public release or on-request delivery arrangement for these artifacts is asserted.

**CRATON evidence.** The technical annex E17–E20 identifies retained model, source, tokenizer, store and evaluation references with their distinct provenance limits [Ozturk, 2026]. The original growth outputs and original 8,000-token packing ledger are not recovered. The separate 25 October packing ledger is retained at RILL 8c400fcd30459b9ff0c5a2aae9c268d4a54eccb1; it is a later reproduction, not the original execution. Private files, external storage and archival objects are not a public reproduction package. No unrestricted checkpoint or corpus release is promised.

**Supplementary observations.** The probe observation is supported by a dated minimal export that omits payload and receipt-body material. The source-change study retains an incomplete acceptance record with local-tree and deployment-identity gaps. These artifacts do not supply missing restoration, isolation or bounded-response evaluations.

Public inspection of a specification is not independent reproduction of the implementation results. Until the corresponding evidence and execution information are made available under an explicit access arrangement, the results remain only partially auditable from the public record.

## 12. Conclusion

Accountable use of AI output requires a reviewer to examine what was recorded as its basis, whether that basis was appropriate and authorized, and whether it remains available for checking. This paper distinguishes claim support, evidence visibility, evidence recoverability, knowledge-boundary isolation and authority fidelity rather than reducing them to answer quality.

We introduced KLM as a proposed open, implementation-neutral specification for an inspectable inference record. The empirical material comprises separate internal experiments rather than one accepted end-to-end deployment. The historical SB-2 summary supports 137/150 platform-domain classifications across 15 classes; the numerator is recovered from the rounded summary, and the result establishes neither source attribution nor correctness of a delivered record. The weight-path study reports 0/15 on its bounded corrected task. The capacity-growth comparison is limited by legacy scoring, missing original outputs and training defects; later corrected experiments do not replace it. Overlap-based scoring fails its silent-versus-contradicted K3 criterion, and the separate detector fails to transfer at its registered live-answer gates. Packing checks expose losses between selection and final-input presence under both stated matching rules. Record checks include adverse production-view outcomes, not only passing samples. These findings motivate separating measurements and retaining negative results. They do not establish that capacity was excluded as a bottleneck, that evidence visibility is complete, or that proposed restoration, boundary-isolation and bounded-response protocols have been executed.

The contribution remains bounded. A conformant-looking record, signature or epistemic label does not establish truth, causal reliance, security or regulatory compliance. Independent implementations, external evaluation, completed protocol tests and a clearly accessible evidence package remain future work. Records can support review of a specific answer, but their usefulness depends on the disclosed profile, retained artifacts and verification inputs; accountability remains with the people and organizations using the system.

## Appendix A. Proposed Evaluation Protocols

This appendix specifies proposed evaluations for RQ4 and RQ5. No completed results under these protocols are reported. The procedures below define intended tests and reporting requirements, not capabilities demonstrated by the implementation. Any later execution must identify its frozen protocol version, populations, configuration and deviations separately from the historical results in Sections 8 and 9.

### A.1 Knowledge-boundary isolation

The proposed threat model evaluates whether a request authorized for one tenant or scope can cause information outside that boundary to be selected, exposed to the model, disclosed in the answer or misattributed in the record. The adversary may submit queries and, where the scenario permits, provide a document containing instructions. Administrative control of the verifier, server, credentials or stored authorization policy is outside this proposed scenario set. This is a protocol scope, not a claim that those excluded attacks are prevented.

Before execution, the evaluation must freeze the tenant/source arrangement, roles and permissions, policy and data versions, attack templates, model configuration, attempt allocation, scoring rubric and adjudication procedure. Authorized-source controls and inaccessible-source controls must confirm that the instrumentation can distinguish permitted, denied and unassessable outcomes. Attempt counts must be fixed in the execution plan; no completed evaluation population is reported here.

The scenario classes are S1, a cross-tenant or cross-scope query; S2, a revoked or out-of-policy source; S3, conflicting authorized and unauthorized sources; and S4, indirect instructions in an admitted document that attempt to widen retrieval or disclosure scope.

Each attempt is scored separately for V1, unauthorized retrieval/selection; V2, unauthorized material in the final model input; V3, unauthorized disclosure in the output; and V4, a record naming a boundary different from the one that governed the response. A single attempt may incur more than one class, so violation counts must not be summed as independent attempts. V4 requires execution-side reference evidence, not merely a plausible record field.

For scenario s and violation class v, let $N_s$ be all selected attempts and $n_{(s,v)}$ the attempts with an observed violation. Report the observed violation fraction $n_{(s,v)} / N_s$ and the unassessable fraction separately, with reasons and evaluation coverage. Unassessable or missing-log attempts remain in the accounting and are not treated as successful isolation. A zero observed violation rate with incomplete coverage is not a clean acceptance result.

Rates describe the stated controls and scenarios only. They do not establish general resistance to prompt injection, data exfiltration or attacks outside the test, and they do not erase knowledge in the base model's parameters.

### A.2 Bounded response

The proposed evaluation contains cases that should be bounded and controls that should receive a supported answer. The should-bound categories are missing evidence, insufficient evidence, inaccessible evidence, materially conflicting evidence and out-of-scope requests. Controls require sufficient, authorized and consistent evidence. Case construction, category counts, expected behaviors, scoring rules and the independent double-scored subset must be fixed before evaluation; no completed population is reported here.

An appropriately bounded response declines the unsupported part, qualifies the material claim to match the evidence, exposes material conflict or refers the matter for human review, as specified by the case rubric. A control is inappropriately bounded when abstention or excessive qualification prevents delivery of an answer that the available authorized evidence supports. A generic hedge does not excuse unsupported material content.

Report appropriate bounding separately by should-bound category and inappropriate bounding on controls. Let $B_c$ be all evaluated cases in should-bound category c, $b_c$ the appropriately bounded cases, C all evaluated controls and a the inappropriately bounded controls. The corresponding rates are $b_c / B_c$ and $a / C$. Failures and unassessable outputs remain in the outcome accounting and are reported separately; they are not silently excluded.

Label consistency has two separate meanings. Computational consistency asks whether the label replays under the record's declared schema and label_formula. Behavioral consistency asks whether the displayed label and explanation agree with the independently assessed evidence and response behavior. Report these separately. A scenario name does not mechanically force CONTESTED or UNCERTAIN independently of the declared formula, and labels from different formula versions must not be regraded as though they shared one rule.

Human assessors should not receive the system's self-assessment before scoring response behavior. Original and adjudicated judgments are retained, and evaluator agreement is reported before adjudication. These procedures operationalize a limited part of authority fidelity; they do not transfer decision authority or accountability to the model.